%% file: 0-Main.tex
\documentclass[letterpaper, journal]{IEEEtran}
\usepackage{cite}
\usepackage{amsmath,amssymb,amsfonts,amsthm}
\usepackage{algorithmic}
\usepackage{graphicx}
\usepackage{textcomp}
\usepackage{tabularx,booktabs,longtable}
\usepackage{array}
\usepackage[caption=false,font=normalsize,labelfont=sf,textfont=sf]{subfig}
\usepackage{stfloats}
\usepackage{url}
\usepackage{verbatim}
\usepackage{color}
\usepackage[colorlinks=true, citecolor=blue, linkcolor=blue, urlcolor=blue]{hyperref} 
\usepackage{orcidlink}
\usepackage{overpic}
\usepackage{textcomp}
\usepackage{anyfontsize}
\usepackage{multirow}
\usepackage[original]{paper-style}

\DeclareMathOperator*{\argmin}{argmin}

\def\BibTeX{{\rm B\kern-.05em{\sc i\kern-.025em b}\kern-.08em
    T\kern-.1667em\lower.7ex\hbox{E}\kern-.125emX}}

\usepackage{balance}
\begin{document}
\title{Communication-Aware Robot Execution for Cloud Inference under Spatially Heterogeneous Connectivity}
\author{
	\vskip 1em
	
	Fengkai Liu\orcidlink{0009-0002-1904-767X}, \emph{Student Member, IEEE},
	Yuichi Ohsita\orcidlink{0000-0001-9784-627X}, \emph{Member, IEEE},
	\\ Masayuki Murata\orcidlink{0000-0002-4168-2875}, \emph{Life Member, IEEE}, 
    and Hideyuki Shimonishi\orcidlink{0009-0004-4364-5877}, \emph{Member, IEEE}

	\thanks{
        This work was supported by JSPS KAKENHI, Grant Number JP25H01118, and by JST SPRING, Grant Number JPMJSP2138.
		
        Fengkai Liu and Masayuki Murata are with Graduate School of Information Science and Technology, The University of Osaka, Osaka, 565-0871, Japan (e-mail: f-liu@ist.osaka-u.ac.jp, murata@ist.osaka-u.ac.jp).
		
        Yuichi Ohsita and Hideyuki Shimonishi is with D3 Center, The University of Osaka, Osaka, 560-0043, Japan (e-mail: yuichi.ohsita.cmc@osaka-u.ac.jp, shimonishi.cmc@osaka-u.ac.jp). 
	}
}


\maketitle

\begin{abstract}
    Cloud-hosted foundation models enable robots to use semantic reasoning beyond onboard computational limits. 
    In this setting, the robot executes a currently available primitive generated by the cloud, and uninterrupted task progress requires the next cloud result before this primitive is exhausted. 
    This execution becomes fragile under spatially heterogeneous connectivity, because the current primitive determines when the next result is needed, whereas the wireless environment determines where the next request can be submitted and where the response can be retrieved. 
    Strategies that reduce latency or improve individual transmissions can shorten this dependency, but they do not determine a submission location that supports reliable upload and leaves a feasible opportunity for response retrieval. 
    To address this problem, we introduce the request--response window, which characterizes the time required for the next cloud cycle, including uplink transmission, cloud inference, downlink retrieval, and inference uncertainty. 
    Building on this window and an available communication map, the proposed framework treats the next request point as a motion decision during ongoing primitive execution, selecting it to provide sufficient communication quality for cloud request submission while preserving progress within the finite support of the current primitive. 
    The selected request point is incorporated into a local planner, which guides the robot toward it before submission and then uses communication and arrival time penalties while the response is pending. 
    Experiments in a simulated indoor wireless scenario with a measurement-based communication map and emulated cloud inference show that the proposed method achieves the highest task success and lowest request failure rates among the compared methods. 
    The request point ablation further supports the contribution of selecting submission locations to execution continuity under spatially varying connectivity. 
\end{abstract}


\input{1-Introduction}
\input{2-Related_Work}
\input{3-Methodology}
\input{4-Evaluation}
\input{5-Conclusion}
\input{6-Appendix}

\bibliographystyle{IEEEtran}
\bibliography{references}



\end{document}

%% file: 1-Introduction.tex
\section{Introduction}
\label{sec_intro}
    Large foundation models are reshaping robot autonomy in open environments. 
    Alongside end-to-end Vision-Language-Action (VLA) policies~\cite{zitkovich23rt2, kim2024openvla, black2026pi0}, modular systems use large language or vision-language models for semantic guidance in tasks such as vision-and-language navigation~\cite{anderson2018vision, zhou2024navgpt, wang2025boosting}, while retaining onboard perception, control, and safety-critical execution~\cite{pan2026robot}. 
    When onboard compute is limited, the models providing this guidance can be hosted in the cloud~\cite{firoozi2024foundation, zheng2025edge}. 
    In modular systems, semantic guidance is translated into executable goals or policies~\cite{shah2022lmnav, liang2023code, huang2023visual, obata2025lipllm, liu2026event}. 
    We consider cloud-guided execution in which each semantic primitive is grounded locally into a motion target. 
    Since the primitive provides finite execution support, uninterrupted progress requires the next primitive before the current one is exhausted. 
    The robot executes the current primitive while the next cloud request may be processed asynchronously, as in related work on asynchronous inference and robot execution~\cite{tang2025vlash, hirose2026asyncvla, nguyen2026speculative}, but continuity still depends on timely completion of the request--response cycle. 
    Under unstable connectivity, this dependency becomes fragile~\cite{pan2026robot, zeydan2025role}: task execution determines when the next result is needed, while network availability varies with the robot's location. 
    
\begin{figure}[!ht]
        \centering
        \includegraphics[width=\linewidth]{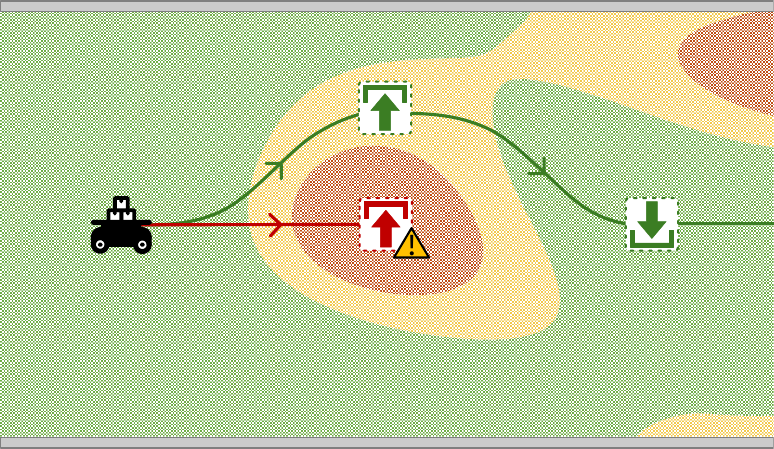}
        \caption{Motivation of communication-aware request placement. Spatially varying connectivity can make a nominal request location unsuitable for submission or response retrieval. Request point selection seeks a favorable submission location that leaves a downstream opportunity to retrieve the response within the current primitive's execution support. }
        \label{fig_motivation}
    \end{figure}

    Existing mitigation strategies do not fully resolve this problem. 
    Methods such as allocating or replicating cloud service resources and using redundant network paths can reduce average latency or improve the reliability of individual transmissions~\cite{ichnowski2023fogros2, chen2024fogros2ls, chen2024fogros2ft, chen2025fogros2plr}. 
    However, they do not determine where the robot should move under spatially varying connectivity. 
    When cloud inference takes long enough for the robot to move substantially before the response returns, the robot's locations at both request submission and response retrieval become critical. 
    The robot may still fail if either request submission or response retrieval occurs in a weak-connectivity region, as illustrated in \PaperFigure~\ref{fig_motivation}. 
    This problem therefore couples robot motion, spatially varying connectivity, and cloud inference latency during execution. 
    The key requirement is to reach a suitable request point and retain a connected retrieval opportunity before the current primitive is exhausted. 
        
    To address this issue, we propose a communication-aware execution framework that selects request points by accounting for cloud inference and transmission delays within the current primitive's finite execution support. 
    The framework first estimates the time required for the next cloud request--response cycle, then uses this estimate to choose a suitable request point and guide the robot toward it before submission. 
    Specifically, the request--response window denotes the estimated duration from request submission to response retrieval, covering network transfer and cloud inference time estimated from recent observations. 
    Using this window, the planner selects a request point, defined as the position from which the next cloud request is submitted. 
    The selected point is chosen to preserve progress toward the current target, avoid excessive detour, and retain a downstream opportunity for response retrieval.
    The request point is incorporated into a Model Predictive Path Integral (MPPI) local planner as a temporary local target before request submission. 
    After submission, the planner resumes tracking the current primitive target while activating communication-aware costs until the next result is buffered. 
    These costs discourage trajectories that would move into weak connectivity regions or exhaust the current primitive too early. 
    Thus, MPPI shapes motion so that task progress, request submission, and response retrieval remain compatible during execution. 

    We evaluate the framework in simulation using an indoor communication map built from measured signal strengths and emulated cloud inference.
    The proposed method achieves the highest task success and lowest request failure rates among the evaluated methods, with fewer request attempts than periodic scheduling. 
    These results support request placement that accounts for spatial connectivity and the current primitive's remaining execution support, beyond asynchronous timing or repeated polling alone. 
    
    The main contributions of this work are as follows: 
    \begin{itemize}
        \item We formulate request point selection for cloud-guided robot execution under spatially heterogeneous connectivity, using the request--response window to relate cloud inference and transmission delays to the current primitive's finite execution support. 
        \item We develop a communication-aware execution framework that coordinates request point selection with local motion planning to support request submission and subsequent response retrieval while preserving task progress. 
        \item We evaluate the framework in a simulated indoor wireless scenario constructed from measured signal-strength data. The comparison and request point ablation support the contribution of request placement to task completion and request reliability. 
    \end{itemize}
    

%% file: 2-Related_Work.tex
\section{Related Work}
    \label{sec_related_work}
    Foundation models have been widely explored for semantic understanding, task planning, and action generation in open robotic environments~\cite{zitkovich23rt2, kim2024openvla, black2026pi0, liang2023code, huang2023visual, obata2025lipllm, wang2025boosting, tariq2025robust, liu2026event, kim2026comp}. 
    Modular robotic systems translate high-level task instructions into navigation goals or executable manipulation policies through separate planning and control components~\cite{shah2022lmnav, liang2023code, yun2025smartlab}. 
    For mobile robots that rely on cloud-hosted models, execution performance is also affected by the communication process that connects local motion with remote inference. 
    We therefore review two closely related directions: latency mitigation through computation and offloading and communication-aware motion planning. 

    \subsection{Latency Mitigation through Computation and Offloading}
        To address the latency in cloud/edge execution pipelines, existing cloud/edge robotics studies have mainly mitigated the problem from system- and execution-level perspectives.
        One line of work focuses on the deployment, configuration, and reliability of remote computing services, reducing the latency of remote module invocation through computation migration, cloud resource selection, service replication, and network interface selection~\cite{ichnowski2023fogros2, chen2024fogros2ls, chen2024fogros2ft, chen2025fogros2plr}. 
        Another line of work directly studies Large Language Model (LLM) inference offloading, cloud-edge collaborative inference, and resource allocation. 
        Existing studies assign LLM requests to cloud or edge servers while jointly considering bandwidth, computation resources, GPU memory, inference latency, and output quality~\cite{he2024large}, or further study edge--cloud partitioned inference for VLA models~\cite{zheng2026rapid}. 
        Subsequent work further summarizes communication-computation coordination problems in LLM inference, training, caching, and transmission from the perspective of mobile edge intelligence~\cite{qu2025mobile}, or selects suitable collaborative inference paths among heterogeneous edge experts, local models, and cloud models to trade off quality, latency, energy consumption, and usage cost~\cite{yuan2025local, jin2026moe2}. 
        In addition, asynchronous inference and robot execution methods attempt to maintain local reactive execution while remote semantic inference is still pending~\cite{hirose2026asyncvla, nguyen2026speculative, yang2026asyncshield}. 
        These studies reduce the impact of remote calls on the execution of robots or mobile terminals, but they mainly address how cloud services are deployed, invoked, scheduled, or latency-masked. 
        
        However, reducing the average latency of remote calls alone is insufficient to guarantee task continuity, because the robot's locations at request submission and response retrieval also affect whether communication is available. 
        Existing latency mitigation and cloud-edge collaboration methods usually treat the request submission location as a consequence of the robot's current state, rather than as a motion decision variable that should be actively optimized. 
        These approaches address latency during execution, but do not directly determine a communication-feasible location for the next cloud request. 

    \subsection{Communication-Aware Motion Planning}
        Another directly related line of research is communication-aware motion planning, which incorporates wireless communication quality into robot motion planning and control. 
        Early work on communication-aware motion planning in mobile networks showed that robot motion affects wireless link quality, and that communication performance can therefore be optimized through trajectory design~\cite{ghaffarkhah2011communication}. 
        Related work also uses the directionality of signal strength to adjust robot positions and provide adaptive communication coverage in multi-robot systems~\cite{gil2015adaptive}. 
        Studies on motion and communication co-optimization further show that path planning, online channel estimation, and communication resources can be incorporated into a unified optimization framework~\cite{yan2013co, muralidharan2017path, ali2019motion, hurst2021communication}. 
        For communication maps and connectivity-region modeling, some methods use wireless measurements or radio maps to discover connectable regions or recover connectivity when link quality degrades~\cite{twigg2013efficient, caccamo2017rcamp}. 
        These studies establish the basic premise of our work: wireless communication quality is not a uniform background condition, but a spatial field that varies with robot location. 
        
        Many communication-aware planning methods further model communication as a persistent path-level or team-level constraint, using measured or predicted signal quality, radio coverage maps, or communication-graph connectivity to maintain link availability or network connectivity~\cite{zavlanos2009distributed, yang2019connectivity, clark2022propeml, yang2024integrating, afzal2023model}. 
        Such formulations emphasize maintaining communication over an extended execution period. 
        Other works relax the requirement of all-time connectivity by planning intermittent communication events and communication locations for multi-robot information exchange~\cite{kantaros2017distributed, kantaros2019temporal}. 
        Edge-assisted navigation also couples motion planning with computation and communication constraints by switching between local and edge planners during execution~\cite{li2025edge}. 
        Related studies use signal maps to guide connectivity-aware route selection~\cite{parwez2026ctmap} and optimize relay stops or transmission points according to link quality and communication requirements~\cite{hurst2023optimization, achey2026rf}. 
        Another related direction optimizes what information should be transmitted under communication constraints, for example by compressing task-relevant map information for online path planning~\cite{psomiadis2024communication}. 
        Together, these works show that communication-aware planning can use robot motion to maintain connectivity, select transmission locations, and support information exchange.

    These two lines of research provide complementary foundations for cloud-guided robot execution: cloud/edge methods address remote computation and response latency, while communication-aware planning accounts for the spatial dependence of wireless communication.
    In the setting considered here, these factors are coupled by the finite execution support of the current primitive.
    While the robot continues moving, the next cloud result must become available before that support is exhausted.
    A suitable request location therefore needs to support submission and leave sufficient time for cloud inference and subsequent response retrieval during ongoing execution.
    Our framework uses the estimated request--response window and the communication map to select such a location while preserving task progress. 


%% file: 3-Methodology.tex
\section{Methodology}
    \label{sec_methodology}
\begin{figure*}[!ht]
        \centering
        {
        \scriptsize
        \definecolor{color1}{RGB}{59,81,188}
        \definecolor{color2}{RGB}{188,59,109}
        \definecolor{color3}{RGB}{111,63,163}
        \newcommand{\ovsym}[2]{
            \makebox(0,0)[l]{\raisebox{-0.5\height}{{\color{#1}#2}}}%
        }
        \begin{overpic}[width=\linewidth,percent]{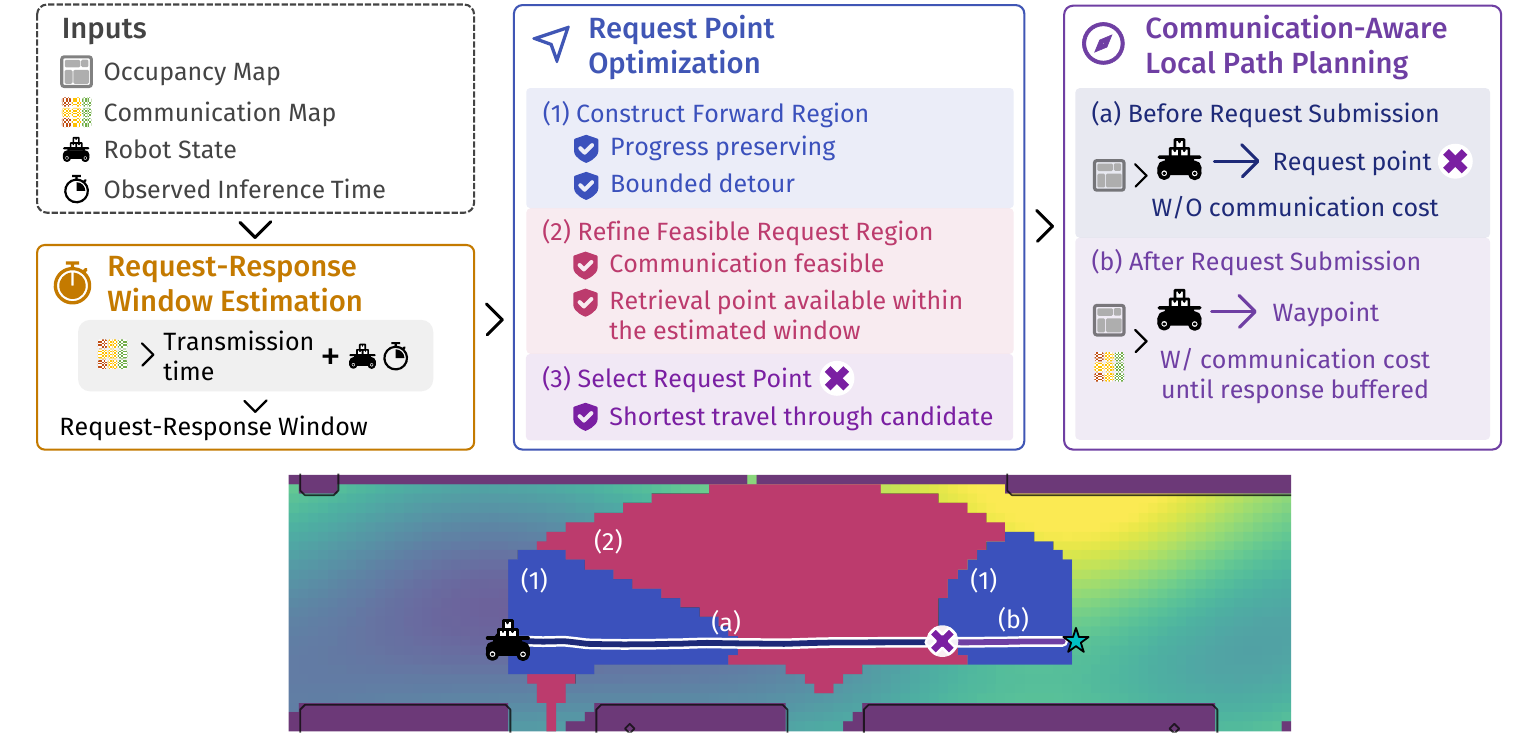}
            \put(14.50,23.25){$T_\text{link}$} 
            \put(24.30,19.75){$\widehat{T}_\text{window}$} 
            \put(56.40,40.60){\ovsym{color1}{$\mathcal{R}_t^{\text{fwd}}$}} 
            \put(60.50,32.85){\ovsym{color2}{$\mathcal{R}_t^{\text{req}}$}} 
            \put(89.80,27.65){\ovsym{color3}{$\mathbf{w}$}} 
        \end{overpic}
        } 
        \caption{System overview. Request--response window estimation combines network transmission time and observed inference latency to estimate the time required for a cloud request--response cycle. Using this window, Request point optimization refines the forward region into the feasible request region and selects a request point that supports submission and preserves a downstream retrieval opportunity. Communication-aware local path planning uses the selected request point as the active target before submission, then switches to the current waypoint with communication-aware soft constraints until the response is buffered.} 
        \label{fig_overview}
    \end{figure*}

    This section presents the proposed communication-aware execution framework, which coordinates local robot motion with cloud requests. 
    We first describe the cloud--local architecture in \PaperSection~\ref{sec_architecture} and formulate the system model in \PaperSection~\ref{sec_problem_formulation}.  
    Based on this formulation, the framework consists of three components, as summarized in \PaperFigure~\ref{fig_overview}:
    \begin{itemize}
        \item \textit{Request--Response Window Estimation} (\PaperSection~\ref{sec_inference_window}), which estimates the time required to complete a request--response cycle from observed inference latency and estimated transmission time. 
        \item \textit{Request Point Optimization} (\PaperSection~\ref{sec_request_trigger}), which computes a suitable request point such that the next cloud request can be submitted from a communication-sufficient region while still preserving a downstream opportunity to retrieve the returned result after passing that point.
        \item \textit{Communication-Aware Local Path Planning} (\PaperSection~\ref{sec_local_planning}), which guides the robot toward the current active target, either the optimized request point or the task primitive target provided by cloud, while incorporating communication-aware soft constraints after the request has been submitted and until the next cloud result has been buffered locally.
    \end{itemize}
    
    \subsection{Cloud--Local Architecture}
        \label{sec_architecture}
        We address the challenge of enabling autonomous robots to execute long-horizon tasks driven by foundation models hosted in cloud, where the separation of cognitive reasoning and physical execution is essential. 
        To support this operation, we adopt a hierarchical control architecture that enforces a strict functional partitioning between the cloud foundation model and the local controller. 
        The cloud foundation model provides semantic guidance. 
        It processes high-dimensional multimodal inputs to interpret user intents and generates abstract task policies rather than precise motor commands. 
        In contrast, the local controller operates as the onboard execution unit responsible for real-time geometric grounding and safety assurance. 
        It maps the cloud-derived semantic intent onto the robot's physical constraints and the local environment to generate feasible trajectories.

        We illustrate this division of labor through navigation and a possible extension to manipulation. 
        In 2D visual navigation, cloud-assisted VLN provides an example: the cloud model uses language instructions and visual observations to identify a target object and specify a coarse destination coordinate or a topological node representing the target semantic region. 
        The local controller plans and executes motion within the traversable space, accounting for local dynamics and geometric and safety constraints not handled by the cloud model. 
        For mobile manipulation, the cloud could specify a target object and manipulation policy, while a task-specific local controller determines feasible base placement and arm motion. 
        The present implementation and evaluation concern planar waypoint navigation. 


    \subsection{Problem Formulation: System Modeling}
        \label{sec_problem_formulation}
        Let $\mathbf{x}(t)$ denote the full system state. 
        For a navigation task, $\mathbf{x}(t)$ reduces to the robot base pose, whereas for a mobile manipulation task it may additionally include manipulator joint states. 
        We denote by $\mathbf{r}(t)=[x(t),\,y(t)]^\top \in \mathbb{R}^2$ the planar position component of the mobile base extracted from $\mathbf{x}(t)$. 
        Let $k$ index the sequence of cloud requests, and let $p(k)$ denote the primitive returned by the $k$-th request. 
        The currently active primitive $p(k)$ denotes the cloud-provided task instruction under execution. 
        Although $p(k)$ may involve the coordination of the mobile base and other actuators, its execution induces a local motion objective for the base. 
        For local planning and execution, we represent this objective by a 2D base waypoint $\mathbf{w}\in\mathbb{R}^2$.
        Depending on the capabilities of the foundation model, $\mathbf{w}$ may be either explicitly provided by the cloud, for example as a navigation coordinate, or implicitly derived by a local kinematic module from a higher-dimensional task goal $\mathbf{g}_{\text{task}}$, for example as a stance pose for object manipulation. 

        In this framework, we assume that the robot operates in a wireless communication environment, such as a 5 GHz Wi-Fi or 5G system, where signal quality varies across space. 
        We further assume the availability of an environmental communication map, denoted as $\mathcal{M}(\mathbf{r})$, which provides an indicator of communication quality at each location, such as Received Signal Strength Indicator (RSSI) in Wi-Fi environments or Reference Signal Received Power (RSRP) in 5G systems. 
        This map allows the robot to query the expected communication condition at any workspace location $\mathbf{r}$ and thereby estimate the transmission cost of a cloud request--response cycle. 

        Based on this system abstraction, the next subsection estimates the time required to complete a cloud request--response cycle.

    \subsection{Request--Response Window Estimation}
        \label{sec_inference_window}
        To maintain seamless autonomy, the robot should obtain the subsequent primitive $p(k+1)$ before the current one can no longer sustain continued task progress. 
        In our setting, this requires the next request to the cloud model to be issued sufficiently early, so that the corresponding request--response cycle can be completed without interrupting ongoing motion. 
        We refer to the time window that must be reserved for this cycle as the request--response window. 
        It spans the interval from request submission to response retrieval, and includes uplink transmission, cloud model inference, downlink retrieval, and an uncertainty margin for inference latency.

        Cloud inference consumes part of the current primitive's remaining execution time while the robot continues moving, and its duration can vary across requests. 
        Assuming the computational load is consistent for similar task types, we estimate the inference latency using an Exponential Weighted Moving Average (EWMA) to smooth fluctuations. 
        The predicted inference time $\widehat{T}_{\text{infer}}(k+1)$ and the exponentially weighted squared prediction error $\sigma_{\text{infer}}^2(k+1)$ are updated recursively as: 
        \begin{align}
            \widehat{T}_{\text{infer}}(k+1) &= \alpha\,T_{\text{infer}}^{\text{obs}}(k) + (1-\alpha)\,\widehat{T}_{\text{infer}}(k) \\
            \sigma_{\text{infer}}^{2}(k+1) &= \alpha\left(T_{\text{infer}}^{\text{obs}}(k) - \widehat{T}_{\text{infer}}(k)\right)^{2} 
            + (1-\alpha)\,\sigma_{\text{infer}}^{2}(k)
        \end{align}
        where $k$ indexes cloud requests and $\alpha \in (0,1)$ is a smoothing factor, $\widehat{T}_{\text{infer}}(0)=T_{\text{infer}}^{\text{obs}}(0)$, $\sigma_{\text{infer}}^{2}(0)=0$.

        The remaining components of the window are the uplink and downlink transmission times. 
        In cloud-based inference, the uplink often carries visual observations for the foundation model, whereas the downlink returns a compact primitive. 
        We therefore first model the uplink transmission, which is the dominant communication component in our setting. 
        The uplink throughput $R_{\text{up}}(\mathbf{r})$ at location $\mathbf{r}$ is estimated by approximating an effective uplink SNR from the environmental communication map.
        Specifically, the effective uplink SNR is estimated from the communication-related signal level at $\mathbf{r}$ provided by the map as
        \begin{align}
            \text{SNR}_{\text{up}}(\mathbf{r}) &= \mathcal{M}(\mathbf{r}) - \Delta P_{\text{tx}} - N_{\text{noise}},
        \end{align}
        where $\Delta P_{\text{tx}}$ is an uplink asymmetry term that represents the effective degradation of the uplink relative to the downlink, and $N_{\text{noise}}$ denotes the noise floor. 
        As the effective SNR decreases, the link becomes more error-prone. 
        We characterize the Packet Error Rate (PER), denoted as $\rho(\mathbf{r})$, using a logistic waterfall curve that captures the sharp degradation of link reliability near the receiver sensitivity threshold $\text{SNR}_{\text{th}}$:
        \begin{align}
            \rho(\mathbf{r}) &= \frac{1}{1 + e^{\kappa \cdot (\text{SNR}_{\text{up}}(\mathbf{r}) - \text{SNR}_{\text{th}})}},
        \end{align}
        where $\kappa$ determines the steepness of the transition. 
        The resulting expected goodput $R_{\text{up}}(\mathbf{r})$ is modeled by accounting for rate adaptation and the retransmission overhead induced by packet errors:
        \begin{align}
            R_{\text{up}}(\mathbf{r}) &= R_{\text{link}}(\mathbf{r}) \cdot (1 - \rho(\mathbf{r}))^{\gamma},
        \end{align}
        where $R_{\text{link}}(\mathbf{r})$ denotes the baseline link rate after rate adaptation at location $\mathbf{r}$, and $\gamma \ge 1$ is the congestion penalty factor. 
        Accordingly, for an uplink payload of size $D_{\text{up}}$, the uplink transmission latency is
        \begin{align}
            T_{\text{up}}(\mathbf{r}) = \frac{D_{\text{up}}}{R_{\text{up}}(\mathbf{r})}.
        \end{align}
        The downlink latency $T_{\text{down}}(\mathbf{r})$ is computed analogously using the downlink payload size $D_{\text{down}}$.

        Finally, we use the prediction error scale to add an uncertainty allowance to the estimated inference time. 
        With an uncertainty weight $\beta$, the request--response window is estimated as
        \begin{align}
            \widehat{T}_{\text{window}}(k+1, \mathbf{r}) =\;& T_{\text{up}}(\mathbf{r}) + T_{\text{down}}(\mathbf{r}) \nonumber \\
            &+ \widehat{T}_{\text{infer}}(k+1) + \beta\,\sigma_{\text{infer}}(k+1),
        \end{align}
        where $\beta\,\sigma_{\text{infer}}(k+1)$ provides an adaptive timing allowance.
        For online window estimation, both transmission terms are evaluated at the candidate submission location $\mathbf{r}$. 
        This single-location approximation keeps the estimation tractable and is motivated by the payload asymmetry discussed above, with a compact returned primitive compared with the uploaded observation. 
        The downlink term therefore approximates transmission at the eventual retrieval location, which may differ from $\mathbf{r}$. 
        The request point optimization uses the estimated window to check for a downstream location satisfying both the retrieval threshold and the timing condition, identifying a retrieval opportunity during continued execution. 

    \subsection{Request Point Optimization}
        \label{sec_request_trigger}
        The request--response window estimated in \PaperSection~\ref{sec_inference_window} specifies how early the next cloud request should be submitted. 
        The planner then uses this estimate together with the communication map to select where the next request should be submitted. 
        The objective is to submit the request from a communication-favorable region while preserving a downstream opportunity to receive the returned primitive during motion toward the current waypoint. 
        Accordingly, once the current waypoint $\mathbf{w}$ becomes available, the planner computes an optimized request point $\mathbf{p}^*$ for the subsequent cloud request.
        The robot first moves toward $\mathbf{p}^*$ and issues the next request upon reaching it. 
        This search runs asynchronously while local control continues toward the waypoint. 
        It then continues toward the current waypoint $\mathbf{w}$ while waiting for the subsequent cloud response. 
        In this way, execution of the current primitive is coupled with preparation for the next cloud request. 

        To compute such a request point, we first consider candidate submission locations along the remaining motion toward $\mathbf{w}$. 
        We formulate this selection through a sequence of candidate filters. 
        Let $\mathcal{C}_{\text{free}}\subset\mathbb{R}^2$ denote the collision-free workspace for the mobile base, accounting for the robot footprint and a safety margin, and let $\tau_{\text{move}}(\mathbf{a},\mathbf{b})$ denote the estimated travel time from $\mathbf{a}$ to $\mathbf{b}$. 
        We estimate the travel time as the Euclidean distance divided by a reference speed, adding a bounded heading-alignment penalty when the initial heading is known. 
        First, to exclude points that lie behind the robot or induce excessive detours, we define the forward region as 
        \begin{align}
            &\mathcal{R}_t^{\text{fwd}} = \Big\{\mathbf{p}\in\mathcal{C}_{\text{free}}\;\Big|\;
            \tau_{\text{move}}(\mathbf{p},\mathbf{w}) < \tau_{\text{move}}(\mathbf{r}(t),\mathbf{w}), \nonumber\\
            &\tau_{\text{move}}(\mathbf{r}(t),\mathbf{p}) + \tau_{\text{move}}(\mathbf{p},\mathbf{w}) \le \tau_{\text{move}}(\mathbf{r}(t),\mathbf{w}) + \varepsilon_{\text{det}}\Big\},
            \label{eq_forward_region}
        \end{align}
        where $\varepsilon_{\text{det}}\ge 0$ is the allowed detour margin.  
        The first condition requires the candidate point $\mathbf{p}$ to be closer to the current waypoint than the robot's present position, so that it lies in the direction of ongoing task progress.
        The second condition bounds the additional detour incurred by visiting $\mathbf{p}$ before resuming motion toward $\mathbf{w}$.

        Since the forward region only enforces task progress and bounded detour, we further check whether each candidate preserves a downstream retrieval opportunity after request submission. 
        For any $\mathbf{p}\in\mathcal{R}_t^{\text{fwd}}$, we define its downstream retrieval margin as
        \begin{align}
            \Delta_t&^{\text{ret}}(\mathbf{p})  = \max_{\mathbf{q}\in\mathcal{R}_t^{\text{fwd}}} \Big\{
            \tau_{\text{move}}(\mathbf{p},\mathbf{q}) - \widehat{T}_{\text{window}}(k+1,\mathbf{p}) \;\Big|\; \nonumber \\ 
            & \tau_{\text{move}}(\mathbf{q},\mathbf{w}) < \tau_{\text{move}}(\mathbf{p},\mathbf{w}), \;\mathcal{M}(\mathbf{q}) \ge S_{\text{ret}} \Big\}.
            \label{eq_retrieval_margin}
        \end{align}
        If no such $\mathbf{q}$ exists, we set $\Delta_t^{\text{ret}}(\mathbf{p})=-\infty$.
        This margin is the largest slack between the time required to move from $\mathbf{p}$ to a downstream communication-feasible point $\mathbf{q}$ and the estimated request--response window for obtaining $p(k+1)$. 
        Thus, a positive margin indicates that at least one downstream point can serve as a feasible retrieval opportunity under the estimated travel time and request--response window. 
        The point $\mathbf{q}$ identifies a possible retrieval location under the travel time estimate. 
        After submission, MPPI guides the robot toward the current waypoint, with the communication cost encouraging motion through communication-favorable regions. 
        
        Based on this quantity, the feasible request region is defined as
        \begin{align}
            \mathcal{R}_t^{\text{req}} = \left\{\mathbf{p}\in\mathcal{R}_t^{\text{fwd}} \;\Big|\;
            \mathcal{M}(\mathbf{p}) \ge S_{\text{sub}}, \Delta_t^{\text{ret}}(\mathbf{p}) \ge \varepsilon_{\text{safe}}\right\},
            \label{eq_request_region}
        \end{align}
        where $S_{\text{sub}}$ and $S_{\text{ret}}$ denote the minimum communication thresholds required for request submission and response retrieval, respectively, and $\varepsilon_{\text{safe}}\ge 0$ is the minimum safety margin.
        Under the adopted window estimate, this region retains candidate submission points that satisfy the request transmission condition and still leave enough time to reach a downstream communication-feasible point for response retrieval. 
        For a fixed robot state and communication map, a longer predicted inference time reduces the downstream retrieval margin and can exclude otherwise communication-feasible request points. 
        
        Finally, the request point is selected from the feasible request region by minimizing the motion cost of reaching the current waypoint through the candidate point: 
        \begin{align}
            \mathbf{p}^* = \argmin_{\mathbf{p}\in\mathcal{R}_t^{\text{req}}} \bigl(\tau_{\text{move}}(\mathbf{r}(t),\mathbf{p}) + \tau_{\text{move}}(\mathbf{p},\mathbf{w})\bigr).
            \label{eq_request_frontier}
        \end{align}
        This criterion selects a communication- and timing-feasible request point that introduces the smallest additional motion cost with respect to the current waypoint.
        Consequently, the robot is guided toward a request point only when it is compatible with the ongoing task motion, rather than being drawn to an isolated communication-favorable location.
        The resulting behavior preserves task progress while avoiding unnecessary detours for request submission.
        
        If $\mathcal{R}_t^{\text{req}}=\varnothing$, no request point is committed at the current cycle. 
        The robot instead continues executing toward $\mathbf{w}$ and retries request point selection after a cooldown. 
        Once the region becomes nonempty, its request point is adopted as the new $\mathbf{p}^*$. 
        If a feasible request point still cannot be obtained after three consecutive recomputations, the planner falls back to submitting the request immediately at the next communication-favorable position encountered along the current motion. 
    
    \subsection{Communication-Aware Local Path Planning}
        \label{sec_local_planning}
        The local planner generates motion while respecting robot dynamics, collision avoidance, and communication requirements. 
        Its nominal task objective remains the current waypoint. 
        Before submission, a selected request point replaces the current waypoint as the active local goal. 
        After submission, the waypoint becomes the active goal again, and the auxiliary costs shape motion while the response is pending. 
        To achieve this, we formulate local motion generation as a constrained optimization and solve it using MPPI control. 
        A sampling based formulation is particularly suitable in our setting because both the obstacle constrained feasible set and the spatial communication field may be highly irregular. 

        At each control time $t$, MPPI rolls out predicted trajectories from the current system state $\mathbf{x}(t)$. 
        Let $\ell=0,\dots,H$ denote the prediction step index within the local horizon $H$. 
        For a predicted trajectory, we write
        \begin{align}
            \mathbf{x}_{\ell} = [x_{\ell},\, y_{\ell},\, \theta_{\ell}]^\top,
            \qquad
            \mathbf{u}_{\ell} = [v_{\ell},\, \omega_{\ell}]^\top ,
        \end{align}
        and $\mathbf{r}_{\ell}=[x_{\ell},\,y_{\ell}]^\top$ denotes the predicted planar position.
        Let $\xi=(\mathbf{x}_{0:H},\mathbf{u}_{0:H-1})$ denote a predicted trajectory and its control sequence, and let $\mathbf{g}(t)$ denote the active local goal used by the planner. 
        Before request submission, $\mathbf{g}(t)$ may be the optimized request point $\mathbf{p}^*(k+1)$; after submission, it returns to the current waypoint $\mathbf{w}(k)$ while the robot waits for $p(k+1)$.
        Consequently, the planner optimizes a control sequence over a horizon of $H$ steps:
        \begin{align}
            \min_{\mathbf{u}_{0:H-1}} ~ & \lambda_{n}J_{\text{nav}}\left(\xi;\mathbf{g}\left(t\right)\right) + \lambda_u J_{\text{ctrl}}(\xi) + \alpha_t \cdot J_{\text{ca}}(\xi) \nonumber \\
            \text{s.t.}\quad 
            & \mathbf{x}_{\ell+1}=f(\mathbf{x}_{\ell},\mathbf{u}_{\ell}), \quad \ell=0,\dots,H-1, \nonumber \\
            & \mathbf{r}_{\ell}\in \mathcal{C}_{\text{free}}, \quad {\ell}=1,\dots,H, \nonumber \\
            & \mathbf{u}_{\ell} \in \mathcal{U}, \quad {\ell}=0,\dots,H-1.
            \label{eq_local_opt}
        \end{align}
        Here, $f(\cdot)$ denotes the robot kinematic model, and $\mathcal{U}$ denotes the admissible control set with $|v|\le v_{\text{max}}$ and $|\omega|\le\omega_{\text{max}}$. 
        The gate $\alpha_t \in \{0,1\}$ controls whether the auxiliary cost $J_{\text{ca}}$ is included in the local rollout objective. 
        It is set to 1 only after the next cloud request has been issued and remains active until the corresponding response is buffered locally. 
        Outside this interval, $\alpha_t=0$, and the local planner reduces to the nominal navigation and control regularization objective.
        Request point selection is performed outside MPPI, and the selected point enters local planning through the navigation reference. 

        We next specify the individual cost terms that compose the local objective. 
        The active goal $\mathbf{g}(t)$ determines the navigation reference $\mathbf{g}_{\ell}(t)$ used at each prediction step. 
        The navigation objective penalizes deviations from this reference over the horizon:
        \begin{align}
            J_{\text{nav}}(\xi;\mathbf{g}(t)) = \sum_{\ell=1}^{H} \lambda_\ell \left\| \mathbf{r}_{\ell}-\mathbf{g}_{\ell}(t) \right\|^2, \quad \lambda_\ell = \frac{\ell}{H}.
            \label{eq_nav_cost}
        \end{align}
        Here, the weights $\lambda_\ell$ increase with the prediction step so that later predicted positions are penalized more strongly. 
        During ordinary waypoint tracking, $\mathbf{g}_{\ell}(t)=\mathbf{w}(k)$ throughout the horizon. 
        During motion toward a request point, the reference remains at $\mathbf{p}^*(k+1)$ through its estimated arrival step, then advances from that point toward the current waypoint at the reference speed. 
        The estimated arrival step is determined by the travel time to the request point at the reference speed, and the reference is recomputed at each control update. 
        
        Next, to improve execution smoothness, we include a small regularization term on control variation:
        \begin{align}
            J_{\text{ctrl}}(\xi) = \frac{1}{H}\sum_{\ell=0}^{H-1} \left\|\mathbf{u}_{\ell}-\mathbf{u}_{\ell-1} \right\|^2 ,
            \label{eq_ctrl_cost}
        \end{align}
        where $\mathbf{u}_{-1}$ is the previously executed control and later differences compare consecutive predicted controls. 
        This term penalizes rapid changes in control commands. 

        Last, after a request has been submitted, and until the corresponding response is buffered locally, the local planner includes communication quality as a soft cost to discourage predicted rollouts from moving into weak-coverage regions. 
        During this interval, it also penalizes rollouts that would reach the current waypoint before the returned primitive becomes available.
        We therefore group the communication penalty and the arrival time penalty into a single communication-aware auxiliary cost:
        \begin{align}
            J_{\text{ca}}(\xi) = \lambda_c J_{\text{comm}}(\xi) + \lambda_a J_{\text{arr}}(\xi),
            \label{eq_ca_cost}
        \end{align}
        where $\lambda_c$ and $\lambda_a$ control the relative importance of maintaining communication quality and delaying premature arrival, respectively. 
        
        We define the communication penalty as
        \begin{align}
            J_{\text{comm}}(\xi) = \frac{1}{H} \sum_{\ell=1}^{H} \left[ \max\!\left( 0,\,  \frac{S_{\text{comm}}-\mathcal{M}(\mathbf{r}_{\ell})}{\Delta_S} \right) \right]^2 ,
            \label{eq_comm_cost}
        \end{align}
        where $S_{\text{comm}}$ is the reference RSSI for the soft communication penalty and $\Delta_S$ is a scaling factor. 
        This reference can exceed the minimum retrieval threshold $S_{\text{ret}}$ to favor stronger coverage. 
        This term is zero in sufficiently strong coverage and increases smoothly when a predicted rollout enters weak-coverage areas.

        After request submission, the robot continues moving toward the current waypoint while waiting for the returned primitive. 
        If a rollout reaches the current waypoint before the response is buffered locally, the executable content of the current primitive is exhausted and the robot has no subsequent target to pursue. 
        This breaks the intended pipelined execution and reduces the system to a stop-and-wait behavior at the waypoint. 
        We therefore introduce an arrival time penalty that discourages such premature arrival. 
        Let $\tau_{\text{arr}}(\xi)$ denote the predicted time to reach the current waypoint.
        For a rollout that reaches the waypoint within the prediction horizon, $\tau_{\text{arr}}(\xi)$ is the first time it enters the arrival tolerance. 
        Otherwise, we estimate it by adding to $H\Delta t$ the remaining travel time from the rollout endpoint, computed using the terminal distance and the rollout's mean forward speed. 
        We define 
        \begin{align}
            J_{\text{arr}}(\xi) = \big[\max\big(0,\, &\widehat{T}_{\text{window}}(k+1,\mathbf{p}^*) \nonumber\\
            &- (t-t_{\text{req}}) - \tau_{\text{arr}}(\xi)\big)\big]^2.
            \label{eq_arr_cost}
        \end{align}
        Here, $t_{\text{req}}$ denotes the time at which the request for $p(k+1)$ is issued. 
        This term becomes positive only when a rollout is predicted to reach $\mathbf{w}$ earlier than the estimated request--response window time, thereby discouraging premature consumption of the current waypoint until the response is buffered locally. 

        Having defined the rollout cost, we now describe how \eqref{eq_local_opt} is solved by MPPI. 
        The MPPI sampling update is well suited to this objective because the feasible set and the communication field can be highly nonconvex. 
        Rollouts that violate the collision constraint 
        $\mathbf{r}_{\ell}^{i} \in \mathcal{C}_{\text{free}}$ 
        or the control bound 
        $\mathbf{u}_{\ell}^{i} \in \mathcal{U}$ 
        are treated as infeasible and excluded from the importance-weighted update. 
        For each feasible rollout, the trajectory score is evaluated using the same objective as in \eqref{eq_local_opt}:
        \begin{align}
            S(\xi^{i})
            =
            \lambda_n J_{\text{nav}}(\xi^{i};\mathbf{g}(t))
            +
            \lambda_u J_{\text{ctrl}}(\xi^{i})
            +
            \alpha_t J_{\text{ca}}(\xi^{i}).
            \label{eq_mppi_score}
        \end{align}
        Let $\mathcal{F}$ denote the set of feasible rollouts. For $\mathcal{F}\ne\varnothing$, their importance weights are
        \begin{align}
            \eta_i =
            \frac{\exp\!\left(-S(\xi^{i})/\lambda\right)}
            {\sum_{j\in\mathcal{F}}\exp\!\left(-S(\xi^{j})/\lambda\right)},
        \end{align}
        where $\lambda$ is the MPPI temperature parameter.
        Let $\mathbf{U}$ denote the nominal control sequence and $\mathbf{U}^{i}$ a feasible sampled control sequence.
        The effective perturbation is $\boldsymbol{\epsilon}^{i}=\mathbf{U}^{i}-\mathbf{U}$.
        We subtract the mean perturbation over feasible rollouts before applying the importance-weighted update:
        \begin{align}
            \overline{\boldsymbol{\epsilon}}
            &= \frac{1}{|\mathcal{F}|}\sum_{j\in\mathcal{F}}\boldsymbol{\epsilon}^{j},\\
            \mathbf{U}^{+}
            &= \Pi_{\mathcal{U}^{H}}\!\left[\mathbf{U}+\sum_{i\in\mathcal{F}}\eta_i\left(\boldsymbol{\epsilon}^{i}-\overline{\boldsymbol{\epsilon}}\right)\right].
        \end{align}
        Here, $\Pi_{\mathcal{U}^{H}}$ projects the updated sequence onto the admissible control set.
        The first updated control is applied to the robot.
        
        Overall, the local MPPI planner shapes motion under dynamics and collision constraints, while request point selection is handled by the preceding optimization stage. 
        

%% file: 4-Evaluation.tex
\section{Evaluation}
\label{sec_evaluation}
This section evaluates whether the proposed communication-aware execution framework improves execution in a measurement-based indoor wireless scenario, without introducing excessive interruption to task execution. 
The main question is whether asynchronous cloud requests can be integrated with robot motion by selecting suitable request locations under spatially heterogeneous connectivity, rather than relying on stopping, periodic polling, or latency compensation alone. 
The remainder of this section describes the experimental setup, baseline methods, parameter settings, and evaluation metrics. 
We then present a qualitative analysis of execution behavior, followed by the main quantitative results, the request point ablation, and discussion. 

\subsection{Experimental Setup}
    \subsubsection{Simulation Environment}
        \label{sec_overview}
        The evaluation is conducted in a ROS-integrated Gazebo simulation environment using a TurtleBot3 Burger model with the platform's kinematic constraints. 
        All methods share the same global task, communication map, and cloud delay model. 
        During execution, the robot queries the cloud-provided communication map to determine whether a request can be submitted or a response retrieved, and to estimate the transmission cost.  

\begin{figure*}[!ht]
            \centering
            \includegraphics[width=\linewidth]{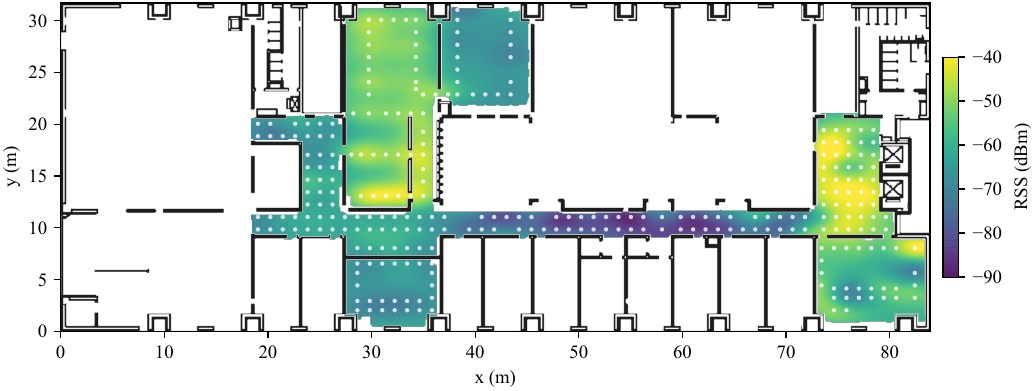}
            \caption{Experimental environment communication map. The map is constructed from the SYL scenario of the SODIndoorLoc dataset~\cite{bi2022supplementary} using RSS fingerprints from MAC208, MAC40, and MAC23, interpolated over the traversable area.}
            \label{fig_env_syl}
        \end{figure*}

        The map is constructed from the SYL scenario of SODIndoorLoc~\cite{bi2022supplementary}, which provides measured WiFi fingerprints and building floor plans. 
        We use fingerprints from MAC208, MAC40, and MAC23 to construct a floor-level communication map that represents a common indoor coverage pattern: office areas are relatively well covered, whereas corridor segments can become weakly connected due to limited nearby access point coverage. 
        For each access point, RSSI measurements are interpolated over the traversable area using Gaussian process regression. 
        Communication quality at each position is the strongest interpolated RSSI among these access points, approximating association to the locally strongest available access point (\PaperFigure~\ref{fig_env_syl}). 
        Planning and simulated communication use this same static RSSI field.
        The task route passes through corridors with weakly connected regions, where request submission becomes infeasible or the estimated transmission cost increases. 
        This measurement-based scenario tests whether the next request can be submitted from a communication-favorable location without substantially disrupting task progress. 

        To remove variability from vision-language model outputs, a mock VLM returns the next waypoint after the simulated request--response delay.
        At each request, the ROS \texttt{move\_base} global planner supplies a path from the robot's current position to the global goal, from which a waypoint is selected at distance $D_{\text{obs}}$ along the path.
        All methods use this waypoint generation rule and the same delay model, while the realized waypoints depend on where requests are made.
        All experiments are executed on the same workstation equipped with an AMD Ryzen 9950X CPU and 64 GB RAM. 

    \subsubsection{Baselines}
        We compare two groups of baselines: request-scheduling methods test when to request the next primitive, while communication-aware path planners test whether incorporating connectivity into the path is sufficient for reliable requests. 
        All baselines share the proposed method's MPPI navigation core, with request point selection and the communication and arrival time penalties disabled. 
        
        \textbf{Request-scheduling baselines.} 
        Stop-and-wait follows the blocking execution pattern considered in~\cite{ruan2026latency}: the robot requests the next primitive after completing the current one and remains stopped until the response arrives. 
        Inference-only triggers the next request based on the estimated inference time $\widehat{T}_{\text{infer}}$ while continuing to execute
        the current primitive. 
        Fixed-period is motivated by asynchronous high-level guidance at a prescribed rate~\cite{wei2026ground}. 
        We evaluate minimum request intervals of $5$, $10$, and $15\,\mathrm{s}$, independent of the remaining support of the current primitive and local communication conditions.
        As in our method, the scheduling baselines allow at most one active request--response cycle. 
        The fixed-period baselines submit the next request as soon as the minimum interval has elapsed and submission is permitted by the active-cycle and retry conditions.
        
        \textbf{Communication-aware path-planning baselines.}
        We adapt CA-RRT*~\cite{hurst2021communication} and MBCAN~\cite{afzal2023model} to the common cloud waypoint interface, denoting the resulting baselines as CA-RRT*-path and MBCAN-path, respectively. 
        Both generate reference paths to the current waypoint for the shared MPPI controller to track. 
        CA-RRT*-path treats connectivity as a soft constraint by penalizing disconnected path length in its fixed-power objective, whereas MBCAN-path imposes a hard RSSI constraint on the reference path, using harmonic-potential navigation within an admissible domain that must also contain the target waypoint. 
        Both use the estimated request--response window to trigger early requests, but not to generate the reference path. 
        
        \textbf{Request point ablation.}
        To isolate the contribution of request point selection, this variant disables request point selection while retaining early submission based on the estimated request--response window and both auxiliary cost terms. 
        Unlike inference-only, which accounts only for inference latency and disables both auxiliary costs, this ablation removes only request point selection. 

    \subsubsection{Parameters and Evaluation Protocol}
        \PaperTable~\ref{tab_parameter_settings} summarizes the main parameters used in the experiments. 
        All baseline methods and the proposed method share the same robot model, motion limits, communication map, cloud delay model, start positions, and global goal.
        We vary the observation range $D_{\text{obs}}$, implemented as waypoint spacing, over $5.0$, $5.5$, and $6.0\,\mathrm{m}$.
        For each value of $D_{\text{obs}}$, we evaluate all methods over 11 start positions, $\{[x,10]\mid x=20,\ldots,30\}$, with the same global goal $[75,10]$.
        Five random seeds are used at each start position, yielding 55 runs per method and observation range and 1,485 runs across the nine main configurations.
        All methods use the same inference duration generator, $\mathcal{N}(6.0,0.5^2)\,\mathrm{s}$. 
        The same five seed values, $0$--$4$, are used for cloud inference and MPPI sampling in each method. 
        Changing the start position shifts where waypoint transitions occur relative to the weakly connected regions. 
        This tests whether each method remains successful when cloud requests are needed at different locations along the same route. 

        The communication model uses an effective uplink receiver sensitivity of $N_{\text{noise}}+\text{SNR}_{\text{th}}=-90\,\mathrm{dBm}$. 
        Its baseline rate is $R_{\text{link}}(\mathbf{r})=1201/(1+\exp[-0.5(\mathcal{M}(\mathbf{r})+75)])\,\mathrm{Mbit/s}$, with RSSI expressed in dBm. 
        The $5\,\mathrm{MB}$ uplink and $0.01\,\mathrm{MB}$ downlink payloads are fixed scenario inputs representing a larger observation and a compact returned primitive. 
        Request candidates are sampled at $0.1\,\mathrm{m}$ spacing, using a $0.22\,\mathrm{m/s}$ reference speed and a heading penalty of $\min(0.20|\Delta\theta|,0.60)\,\mathrm{s}$ when the initial heading is known. 
        Each request has a $15\,\mathrm{s}$ timeout, after which its response is no longer accepted for handoff. 
        Failed requests are retried after a $0.5\,\mathrm{s}$ cooldown when the method's trigger condition is met. 
        If the robot reaches a non-final waypoint without the next primitive, it holds there while the request process continues. 
        Runs end upon reaching the global goal, a terminal planning failure, or external termination at the run time limit. 
        All results use a common $480\,\mathrm{s}$ wall-clock observation window, including simulator startup. 
        Task completion and waypoint waiting times are computed from simulation timestamps. 
        
        For MBCAN-path, the main comparison includes hard-domain thresholds of $-84$ and $-87\,\mathrm{dBm}$. 
        We also evaluate $-90\,\mathrm{dBm}$ to examine whether further relaxing the path constraint improves task completion. 
        All three settings use the same start positions and random seeds at each observation range. 
        The varied threshold defines the admissible path domain and is distinct from the request submission and response retrieval thresholds. 

        \begin{table}[t]
            \centering
            \caption{Experimental parameters.}
            \label{tab_parameter_settings}
            \footnotesize
            \begin{tabular}{clc}
                \toprule
                Symbol & Parameter & Value \\
                \midrule
                $\mathbf{x}(0)$ & Initial robot positions & $[20,10]$--$[30,10]$ \\
                $\mathbf{g}$ & Global goal & $[75.0,\,10.0]$ \\
                $v_{\text{max}}$ & Maximum linear velocity & $0.22\,\mathrm{m/s}$ \\
                $\Delta t$ & Control interval & $0.1\,\mathrm{s}$ \\
                $H$ & MPPI prediction horizon & $30$ \\
                $K$ & Number of MPPI rollouts & $300$ \\
                $\lambda$ & MPPI temperature & $1.0$ \\
                $\lambda_n$ & Navigation cost weight & $1.0$ \\
                $\lambda_u$ & Control regularization weight & $0.005$ \\
                $\lambda_c$ & Communication cost weight & $15.0$ \\
                $\lambda_a$ & Arrival time cost weight & $10.0$ \\
                $S_{\text{sub}}$ & Uplink submission threshold & $-84\,\mathrm{dBm}$ \\
                $S_{\text{ret}}$ & Downlink retrieval threshold & $-90\,\mathrm{dBm}$ \\
                $S_{\text{comm}}$ & Communication penalty reference & $-84\,\mathrm{dBm}$ \\
                $\Delta_S$ & Communication cost scale & $10.0\,\mathrm{dB}$ \\
                $\kappa$ & PER transition steepness & $2.0$ \\
                $\gamma$ & Congestion penalty factor & $1.5$ \\
                --- & Base delay per transmission & $0.05\,\mathrm{s}$ \\
                $D_{\text{up}}$ & Uplink payload size & $5.0\,\mathrm{MB}$ \\
                $D_{\text{down}}$ & Downlink payload size & $0.01\,\mathrm{MB}$ \\
                $D_{\text{obs}}$ & Waypoint spacing & $\{5.0,\,5.5,\,6.0\}\,\mathrm{m}$ \\
                $T_{\text{infer}}$ & Inference duration generation & $\mathcal{N}(6.0,\,0.5^2)\,\mathrm{s}$ \\
                $\Delta P_{\text{tx}}$ & Uplink asymmetry offset & $6.0\,\mathrm{dB}$ \\
                $\alpha$ & EWMA smoothing factor & $0.3$ \\
                $\beta$ & Inference uncertainty coefficient & $1.0$ \\
                $\widehat{T}_{\text{infer}}(0)$ & Initial inference time estimate & $6.0\,\mathrm{s}$ \\
                $\varepsilon_{\text{det}}$ & Detour margin & $4\,\mathrm{s}$ \\
                $\varepsilon_{\text{safe}}$ & Retrieval safety margin & $0.5\,\mathrm{s}$ \\
                \bottomrule
            \end{tabular}
        \end{table}
        
    \subsubsection{Evaluation Metrics}
        We evaluate each method by measuring task completion and request behavior during robot execution. 
        The task success rate $R_{\text{succ}}$ is the percentage of runs in which the robot reaches the global goal $\mathbf{g}$. 
        The number of cloud requests $N_{\text{req}}$ counts recorded request attempts after the initial waypoint becomes available, and the request failure rate $R_{\text{fail}}$ is the fraction of those attempts recorded as failed. 
        Failures include unsuccessful cloud calls, request timeouts, and requests still unresolved when the observation window ends. 
        The waiting time $T_{\text{wait}}$ is the cumulative time spent holding at non-final waypoints while the next primitive is unavailable, measured from waypoint arrival until the next waypoint is received or the run terminates. 
        We report the simulated inference duration $T_{\text{infer}}$ as a timing reference, and $T_{\text{task}}$ as the execution duration of successful runs. 
        Each method is evaluated over 55 runs per observation range. 
        Except for $R_{\text{succ}}$, the main table reports means and sample standard deviations across runs, including unsuccessful tasks. 
        Inference durations are averaged over successful requests within each primitive and then over primitives within a run, excluding runs with no valid duration samples. 
        The completion time figure shows each method's successful runs and per-start medians. 

\subsection{Qualitative Analysis}
\begin{figure*}[!ht]
        \centering
        \includegraphics[width=\textwidth]{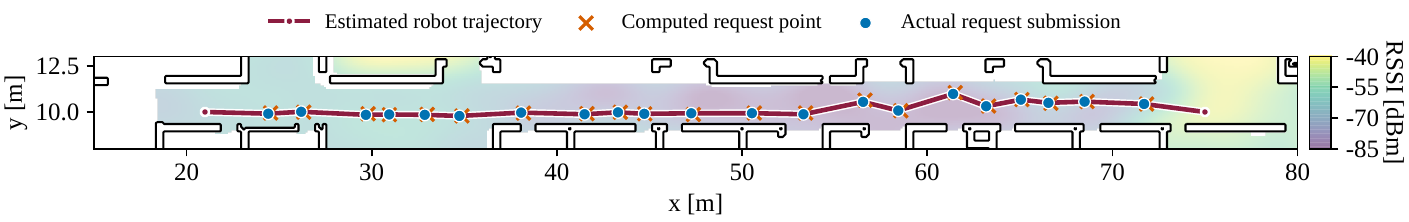}
        \caption{Illustrative execution trace from start (21, 10) m at $D_{\text{obs}}=5.0\,\mathrm{m}$, showing computed request points and 21 actual submissions. The trace connects recorded positions. }
        \label{fig_case_study}
    \end{figure*}

    \PaperFigure~\ref{fig_case_study} illustrates how our method adjusts request locations to the spatial variation in connectivity. 
    In the weakly connected corridor around $x=58$--$63\,\mathrm{m}$, the RSSI at a cloud-generated waypoint falls below the submission threshold. 
    The robot submits the next request from a connected location before reaching this waypoint and receives the response while still moving toward it. 
    Thus, the weak connectivity at the waypoint does not force the robot to stop there and attempt an upload. 

    As the task guidance advances, the method selects a new request point according to the updated robot state, waypoint, and communication map. 
    The upward deviation near $x=61\,\mathrm{m}$ shows a lateral adjustment toward stronger coverage, with the actual submission occurring close to the selected point. 
    The submission satisfies the communication constraint, and the robot continues moving while awaiting the response. 
    Together, these observations show how early submission and lateral repositioning accommodate weakly connected waypoints without requiring the entire route to satisfy the submission threshold. 
    All 21 recorded requests succeed, and the robot completes the task. 

\subsection{Quantitative Analysis}
    \begin{table*}[!ht]
        \centering
        \caption{Overall execution performance under different observation ranges.
            Each row summarizes 55 runs (11 starts and five seeds). 
            Units: $D_{\text{obs}}$ (m), $R_{\text{succ}}$ and $R_{\text{fail}}$ (\%), $T_{\text{wait}}$ and $T_{\text{infer}}$ (s). 
            Except for $R_{\text{succ}}$, entries are mean $\pm$ sample standard deviation.
            The RSSI thresholds in MBCAN-path labels define the admissible path domain. 
            Bold marks the highest $R_{\text{succ}}$ and lowest mean $T_{\text{wait}}$ and $R_{\text{fail}}$ within each observation range, including ties. 
        }
        \label{tab_overall}
        \small
        \setlength{\tabcolsep}{4.2pt}
        \begin{tabular}{@{}clccccc@{}}
            \toprule
            \multicolumn{1}{c}{$D_{\text{obs}}$} & \multicolumn{1}{c}{Method} & \multicolumn{1}{c}{$R_{\text{succ}}$} & \multicolumn{1}{c}{$T_{\text{wait}}$} & \multicolumn{1}{c}{$R_{\text{fail}}$} & \multicolumn{1}{c}{$T_{\text{infer}}$} & \multicolumn{1}{c}{$N_{\text{req}}$} \\ 
            \midrule
            \multirow{9}{*}{5.0} & Stop-and-wait & 41.8 & $187.89 \pm 99.03$ & $43.99 \pm 36.64$ & $5.9 \pm 0.2$ & $19.3 \pm 7.7$ \\
             & Inference-only & 38.2 & $180.80 \pm 135.38$ & $47.19 \pm 32.80$ & $5.9 \pm 0.2$ & $26.5 \pm 9.9$ \\
             & Fixed-period, 5s & 87.3 & $31.08 \pm 73.82$ & $8.29 \pm 12.50$ & $6.0 \pm 0.1$ & $37.0 \pm 9.7$ \\
             & Fixed-period, 10s & 78.2 & $73.15 \pm 118.84$ & $18.76 \pm 23.75$ & $6.0 \pm 0.1$ & $27.1 \pm 8.0$ \\
             & Fixed-period, 15s & 43.6 & $168.83 \pm 138.55$ & $40.04 \pm 29.20$ & $6.0 \pm 0.2$ & $24.7 \pm 7.3$ \\
            \cmidrule(lr){2-7}
             & CA-RRT$^*$-path & 87.3 & $39.16 \pm 94.03$ & $25.37 \pm 21.71$ & $6.0 \pm 0.2$ & $17.9 \pm 6.5$ \\
             & MBCAN-path, $-84$dBm & 0.0 & {\boldmath $0.00 \pm 0.00$} & $6.22 \pm 6.44$ & $5.9 \pm 0.3$ & $6.5 \pm 2.4$ \\
             & MBCAN-path, $-87$dBm & 80.0 & $18.18 \pm 64.75$ & $18.42 \pm 19.42$ & $6.0 \pm 0.2$ & $13.4 \pm 5.7$ \\
            \cmidrule(lr){2-7}
             & Ours & \textbf{94.5} & $5.26 \pm 39.03$ & {\boldmath $0.22 \pm 1.16$} & $6.0 \pm 0.1$ & $18.1 \pm 2.3$ \\
            \midrule
            \multirow{9}{*}{5.5} & Stop-and-wait & 32.7 & $197.56 \pm 106.91$ & $49.46 \pm 37.86$ & $5.9 \pm 0.3$ & $19.6 \pm 8.3$ \\
             & Inference-only & 43.6 & $165.38 \pm 140.44$ & $45.11 \pm 35.32$ & $5.9 \pm 0.2$ & $24.1 \pm 10.3$ \\
             & Fixed-period, 5s & 92.7 & $5.22 \pm 5.32$ & $4.26 \pm 1.77$ & $6.0 \pm 0.1$ & $35.9 \pm 11.0$ \\
             & Fixed-period, 10s & 94.5 & $19.37 \pm 61.66$ & $8.52 \pm 12.50$ & $6.0 \pm 0.1$ & $23.8 \pm 5.3$ \\
             & Fixed-period, 15s & 54.5 & $134.75 \pm 141.94$ & $33.29 \pm 29.92$ & $6.0 \pm 0.2$ & $22.5 \pm 7.6$ \\
            \cmidrule(lr){2-7}
             & CA-RRT$^*$-path & 80.0 & $57.43 \pm 117.40$ & $30.38 \pm 27.09$ & $5.9 \pm 0.2$ & $17.6 \pm 8.2$ \\
             & MBCAN-path, $-84$dBm & 12.7 & $0.28 \pm 0.66$ & $7.23 \pm 7.59$ & $5.9 \pm 0.2$ & $6.5 \pm 2.7$ \\
             & MBCAN-path, $-87$dBm & 87.3 & $1.32 \pm 2.35$ & $17.84 \pm 13.01$ & $5.9 \pm 0.2$ & $11.1 \pm 3.7$ \\
            \cmidrule(lr){2-7}
             & Ours & \textbf{100.0} & {\boldmath $0.13 \pm 0.95$} & {\boldmath $0.69 \pm 2.00$} & $6.0 \pm 0.2$ & $16.6 \pm 1.5$ \\
            \midrule
            \multirow{9}{*}{6.0} & Stop-and-wait & 34.5 & $183.63 \pm 108.62$ & $48.44 \pm 38.49$ & $5.9 \pm 0.3$ & $18.3 \pm 8.8$ \\
             & Inference-only & 47.3 & $154.45 \pm 142.26$ & $43.42 \pm 37.25$ & $5.9 \pm 0.2$ & $22.2 \pm 10.5$ \\
             & Fixed-period, 5s & 85.5 & $3.02 \pm 3.95$ & $4.43 \pm 1.98$ & $6.0 \pm 0.1$ & $37.4 \pm 13.6$ \\
             & Fixed-period, 10s & 98.2 & $3.91 \pm 5.14$ & $6.71 \pm 3.20$ & $6.0 \pm 0.1$ & $22.5 \pm 3.8$ \\
             & Fixed-period, 15s & 67.3 & $90.73 \pm 131.45$ & $24.35 \pm 27.83$ & $6.0 \pm 0.2$ & $20.4 \pm 7.5$ \\
            \cmidrule(lr){2-7}
             & CA-RRT$^*$-path & 94.5 & $16.75 \pm 64.13$ & $22.40 \pm 16.25$ & $5.9 \pm 0.2$ & $13.3 \pm 4.7$ \\
             & MBCAN-path, $-84$dBm & 45.5 & $0.46 \pm 1.52$ & $13.80 \pm 25.45$ & $5.9 \pm 0.2$ & $6.6 \pm 4.0$ \\
             & MBCAN-path, $-87$dBm & 78.2 & $0.37 \pm 0.68$ & $21.65 \pm 21.66$ & $5.9 \pm 0.2$ & $9.2 \pm 4.4$ \\
            \cmidrule(lr){2-7}
             & Ours & \textbf{100.0} & {\boldmath $0.00 \pm 0.00$} & {\boldmath $0.14 \pm 1.04$} & $6.0 \pm 0.2$ & $15.5 \pm 1.7$ \\
            \bottomrule
        \end{tabular}
    \end{table*}
    
    \subsubsection{Overall Performance}
\begin{figure*}[!ht]
            \centering
            \includegraphics[width=\linewidth]{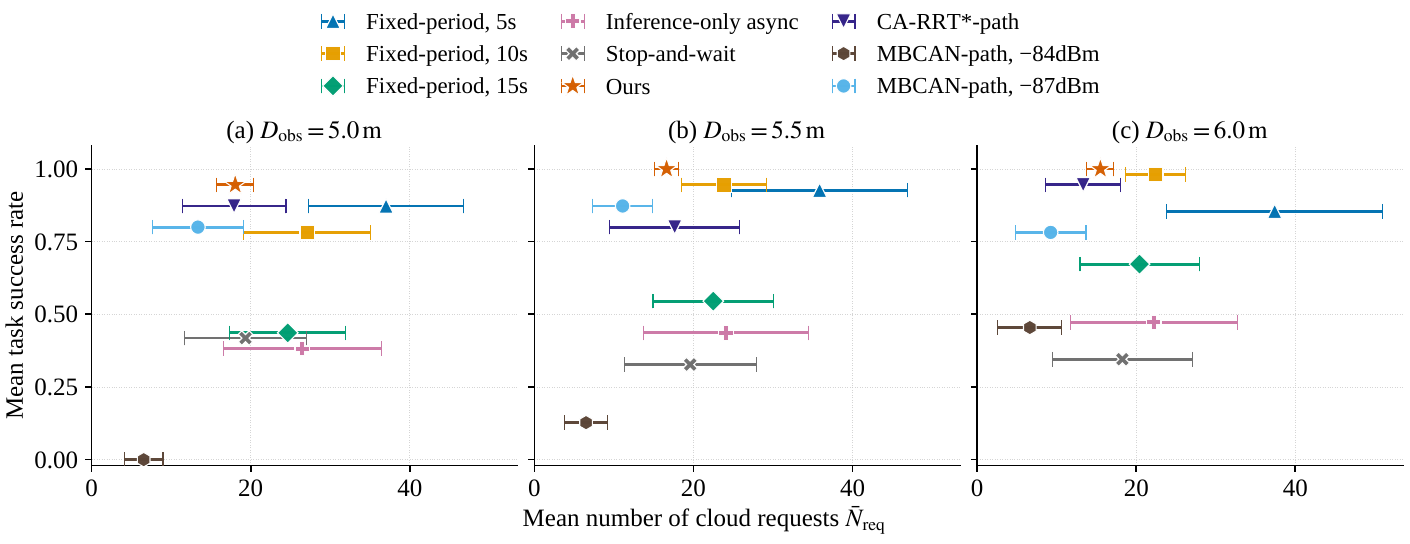}
            \caption{Task success rate versus mean number of cloud requests under different observation ranges. Horizontal error bars indicate one sample standard deviation in request count. Points closer to the upper left indicate higher task success with fewer requests.}
            \label{fig_task_succ_scatter}
        \end{figure*}

        \PaperTable~\ref{tab_overall} and \PaperFigure~\ref{fig_task_succ_scatter} compare task completion and request load across the three observation ranges. 
        Mean inference durations range from 5.90 to $5.99\,\mathrm{s}$ across methods and observation ranges, indicating comparable cloud inference latency across the evaluated configurations. 
        Our method achieves the highest task success rate across all three observation ranges, completing all runs at $D_{\text{obs}}=5.5$ and $6.0\,\mathrm{m}$ and 52 of 55 runs at $5.0\,\mathrm{m}$. 
        Among the request-scheduling baselines, stop-and-wait and inference-only succeed in fewer than half of the runs at each observation range, as neither waiting for a response nor advancing submission based on inference latency avoids unfavorable request locations. 
        Fixed-period execution achieves higher task success through repeated request attempts, but shortening the interval does not consistently improve completion. 
        Reducing the interval from 10 to 5 s increases request load at all three observation ranges, while improving task success only at $D_{\text{obs}}=5.0\,\mathrm{m}$. 
        Our method achieves higher task success with fewer requests than either setting. 
        At $D_{\text{obs}}=6.0\,\mathrm{m}$, for example, it completes all runs with approximately 31.1\% fewer requests than fixed-period 10 s and a mean request failure rate of 0.14\%, compared with 6.71\% for that baseline. 
        These results support the benefit of communication-aware request placement over increasing request frequency alone.

        The communication-aware path-planning baselines improve connectivity along the path, but do not explicitly select where the next cloud request should be submitted.
        For CA-RRT$^*$-path, communication quality enters the path objective as a soft cost, so the selected path may still contain weakly connected segments. 
        Submission can therefore occur at a location that does not satisfy the request submission threshold, consistent with its mean request failure rates of 22.40\%--30.38\%. 
        MBCAN-path instead restricts its reference path to a hard communication domain, which can prevent the robot from reaching a cloud-generated waypoint outside that domain. 
        At $D_{\text{obs}}=5.0\,\mathrm{m}$, the $-84\,\mathrm{dBm}$ setting completes no tasks, with most runs terminating because no admissible path to the returned waypoint is found. 
        Relaxing the domain threshold to $-87\,\mathrm{dBm}$ improves task success across all three observation ranges, but also admits locations below the unchanged request submission threshold. 
        Its mean request failure rates remain between 17.84\% and 21.65\%.
        Our method separates request point selection from waypoint following, allowing submission from a communication-feasible location even when the current waypoint lies in a weakly connected region. 
        This distinction helps explain its higher task success and lower request failure rates. 
        The threshold effect is examined further in the discussion. 
         
        The benefit of request placement is also reflected in reduced waiting for the next primitive. 
        Our method incurs no waypoint waiting in 163 of 165 runs, with mean cumulative waiting times of 5.26, 0.13, and $0.00\,\mathrm{s}$ across the three observation ranges. 
        By comparison, Fixed-period 10 s incurs mean waiting times of 73.15, 19.37, and $3.91\,\mathrm{s}$. 
        Low waiting time must nevertheless be interpreted alongside task completion. 
        For example, MBCAN-path with a $-84\,\mathrm{dBm}$ threshold records zero waiting at $D_{\text{obs}}=5.0\,\mathrm{m}$ but completes no tasks. 
        Most runs terminate because no path to the returned waypoint is found within the admissible domain, before the robot reaches that waypoint and begins waiting. 

    \subsubsection{Ablation Study}
        \begin{table}[!ht]
            \centering
            \caption{Request-placement ablation. 
                The ablation disables request point selection while leaving the remaining components unchanged. 
                Units: $D_{\text{obs}}$ (m), $R_{\text{succ}}$ and $R_{\text{fail}}$ (\%). 
                Bold indicates the higher task success rate and lower mean request failure rate within each observation range. 
            }
            \label{tab_ablation}
            \small
            \setlength{\tabcolsep}{6pt}
            \begin{tabular}{clcc}
                \toprule
                \multicolumn{1}{c}{$D_{\text{obs}}$} & \multicolumn{1}{c}{Configuration} & \multicolumn{1}{c}{$R_{\text{succ}}$} & \multicolumn{1}{c}{$R_{\text{fail}}$} \\
                \midrule
                \multirow{2}{*}{5.0} & Ours & \textbf{94.5} & \textbf{0.22} \\
                 & w/o request placement & 65.5 & 19.16 \\
                \midrule
                \multirow{2}{*}{5.5} & Ours & \textbf{100.0} & \textbf{0.69} \\
                 & w/o request placement & 78.2 & 17.19 \\
                \midrule
                \multirow{2}{*}{6.0} & Ours & \textbf{100.0} & \textbf{0.14} \\
                 & w/o request placement & 90.9 & 14.38 \\
                \bottomrule
            \end{tabular}
        \end{table}
    
    \PaperTable~\ref{tab_ablation} examines the contribution of request point selection, with full-window triggering and both auxiliary costs retained in the ablation. 
    Across the three observation ranges, mean request failure rates are $14.38$--$19.16\%$ without request point selection, compared with $0.14$--$0.69\%$ for our method. 
    Full-window triggering advances submission according to the estimated request--response delay, but does not by itself move the robot to a communication-feasible submission location. 
    In the ablation, requests are issued from positions reached while tracking the current waypoint. 
    In a representative execution trace, the robot repeatedly submits requests below the submission threshold without receiving the next waypoint. 
    This example illustrates why accounting for the response delay alone does not ensure submission feasibility: repeated requests can still fail when their locations remain unfavorable. 
    Request point selection instead provides an explicit motion target for submission, allowing the local planner to guide the robot toward a communication-favorable location before issuing the request. 
    The task-success statistics also include local planning failures associated with collision risk near walls. 
    Within the tested framework, these results support the contribution of request point selection to request reliability and overall task completion beyond full-window triggering and the auxiliary costs. 
    The largest gains occur at the shortest observation range. 

\subsubsection{Discussion}
\begin{figure*}[!ht]
        \centering
        \includegraphics[width=\linewidth]{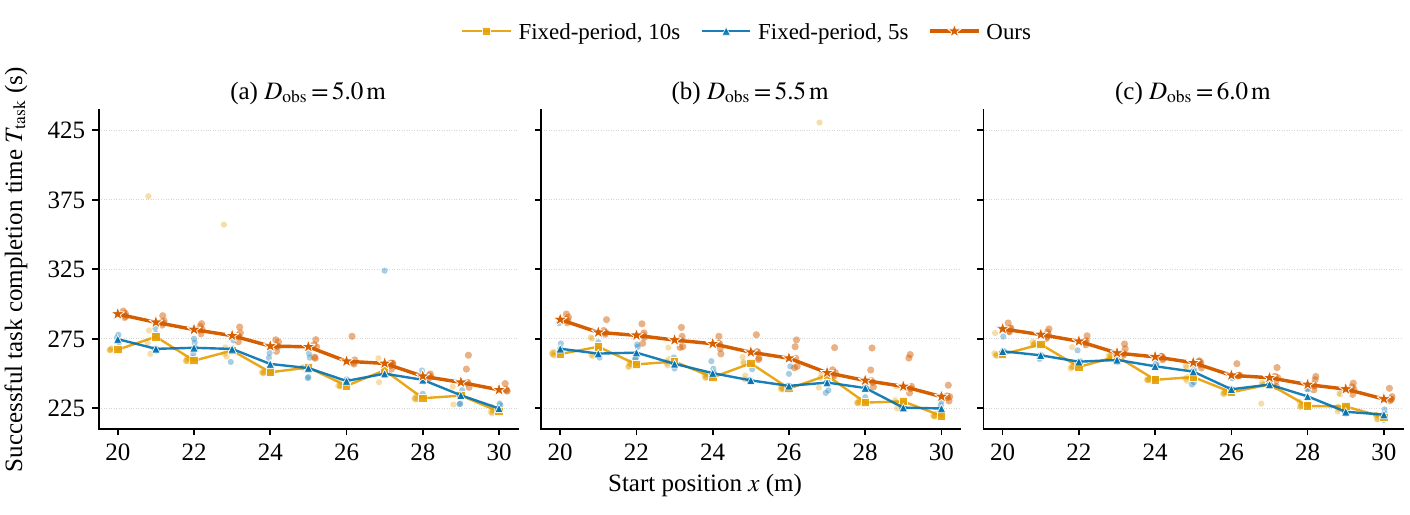}
        \caption{Task completion time across start positions and observation ranges for successful runs. Translucent points show individual runs; connected markers indicate per-start medians. Each method includes its own successful runs, so the plotted samples are not necessarily paired.}
        \label{fig_task_time}
    \end{figure*}

    The experimental results reveal several trade-offs between task completion, execution time, and communication feasibility.
    First, \PaperFigure~\ref{fig_task_time} compares our method with Fixed-period 5 s and 10 s, the two scheduling configurations with the highest task success rates. 
    Together with \PaperTable~\ref{tab_overall}, it shows that our method achieves higher task success with fewer requests while maintaining comparable completion times on successful runs. 
    The lower waypoint waiting time does not, however, consistently lead to shorter execution. 
    As illustrated in \PaperFigure~\ref{fig_case_study}, reaching a communication-favorable request point can require additional travel and turning before submission. 
    These motion costs, together with other differences in local execution, contribute to completion time but are not captured by waypoint waiting. 
    Our method thus reduces interruptions at primitive transitions while allowing additional motion to support reliable request submission. 
    
\begin{figure*}[!ht]
        \centering
        \includegraphics[width=\linewidth]{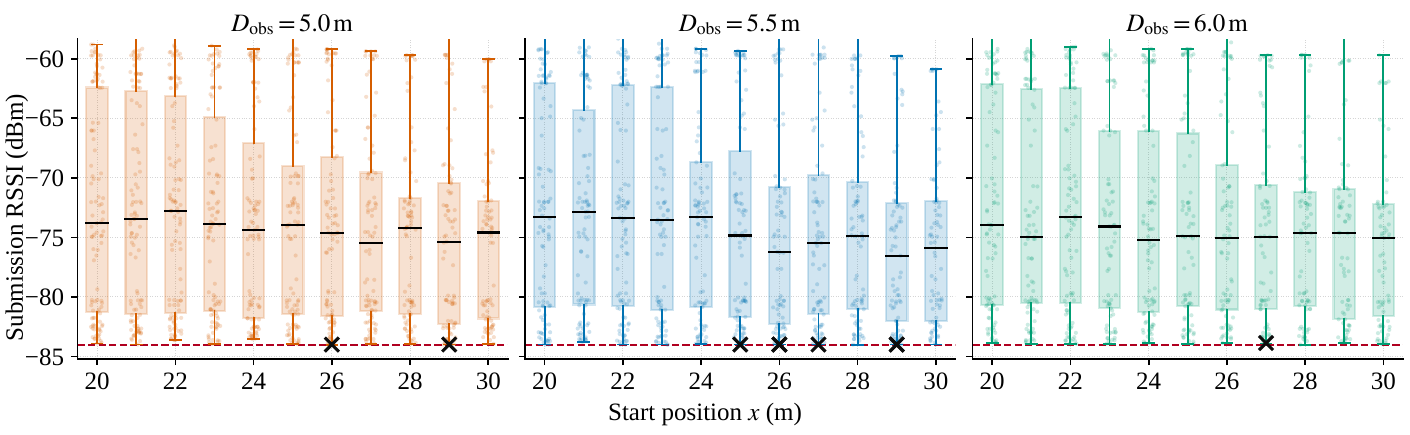}
        \caption{RSSI at our method's request submission positions across observation ranges. Boxes show medians and interquartile ranges, with whiskers extending to the most extreme values within 1.5 interquartile ranges of the box. Translucent points represent individual submissions; black crosses mark requests that subsequently fail. The dashed line denotes the $-84\,\mathrm{dBm}$ submission threshold. }
        \label{fig_rssi_distribution}
    \end{figure*}

    Second, the MBCAN-path results illustrate how the imposed RSSI threshold affects reference-path feasibility and request reliability. 
    For MBCAN-path at $D_{\text{obs}}=5.0\,\mathrm{m}$ with a $-84\,\mathrm{dBm}$ domain threshold, 49 of 55 runs terminate because the planner finds no path within the admissible domain. 
    In 46 of these runs, the returned waypoint lies below the threshold. 
    The hard domain thus excludes waypoints that remain part of the cloud-generated task guidance. 
    Relaxing the threshold to $-87\,\mathrm{dBm}$ improves task completion, but further relaxation to $-90\,\mathrm{dBm}$ provides no additional improvement: successful runs decrease from 44 to 38 at $5.0\,\mathrm{m}$, remain at 48 at $5.5\,\mathrm{m}$, and decrease from 43 to 42 at $6.0\,\mathrm{m}$. 
    Mean request failure rates also increase at all three ranges. 
    Enlarging the admissible domain permits motion through weaker-connectivity regions without changing the submission requirement, so improved path feasibility alone does not ensure reliable requests. 
    Our method instead places the communication constraint on the request location, allowing task waypoints to remain in weakly connected regions.
    \PaperFigure~\ref{fig_rssi_distribution} shows that all recorded submissions satisfy the $-84\,\mathrm{dBm}$ threshold, with median margins of 9.3--9.8 dB across observation ranges.
    The actual submission positions therefore usually provide some margin beyond minimum feasibility.
    Nine requests nevertheless fail after submission, indicating that a feasible submission position does not ensure success of the entire request--response cycle.
    The RSSI measured at submission cannot determine the causes of these later failures. 

    Finally, the results across observation ranges illustrate how the support provided by the current primitive affects execution. 
    Our method completes 52 of 55 runs at $D_{\text{obs}}=5.0\,\mathrm{m}$ and all runs at the two larger settings. 
    This pattern is consistent with larger waypoint spacing allowing more time and a wider forward region for selecting a request point that satisfies the submission and downstream retrieval conditions. 
    In a deployed VLN system, the available support also depends on the waypoint generated from the current observations and the robot's ability to execute it. 
    

%% file: 5-Conclusion.tex
\section{Conclusion}
    \label{sec_conc}
    We studied robot execution based on cloud-hosted foundation model inference under spatially heterogeneous wireless connectivity.
    Task continuity in this setting depends on both the motion supported by the current cloud primitive and the communication conditions encountered when requesting the next one.
    To capture this coupling, we introduced the request--response window and developed a communication-aware execution framework that selects request points from a communication map and incorporates them into local planning.
    This allows the robot to submit requests from communication-favorable locations while continuing toward the task goal, even when cloud-generated waypoints lie in weakly connected regions.
    Experiments in a ROS-integrated simulation with a measurement-based indoor communication map and emulated cloud inference showed higher task success and lower request failure rates than the evaluated baselines.
    Compared with periodic requests, our method also required fewer request attempts and incurred little waypoint waiting, with completion times in a comparable range for successful runs. 
    The ablation further supported the contribution of request point selection beyond full-window triggering and the auxiliary motion costs.
    These results demonstrate the value of coordinating local motion with communication opportunities during cloud-guided execution.

    Several directions remain for future work.
    The current framework assumes that a communication map is available before execution, whereas human movement, occlusion, network load, and environmental changes can alter communication quality in practice.
    Extending request point selection to incomplete or dynamically changing maps would broaden its applicability.
    Experiments with online cloud inference over deployed Wi-Fi or 5G networks are also needed to evaluate the framework under real communication and model-output variability.
    In VLN, such experiments could further examine how request placement interacts with observation-dependent guidance from the cloud model.
    Finally, when no communication-feasible request location is available within the support of the current primitive, local fallback strategies, degraded execution, or task replanning could help maintain task progress.
    

%% file: 6-Appendix.tex
\section*{Appendix}
        \label{sec_capability}
        We examine the request deferral boundary for continuous reference motion at the constant speed $v_{\text{max}}$. 
        Let $\mathbf{p}_k$ denote the observation/submission location associated with request $k$, let $\mathbf{w}(k)$ denote the waypoint induced by the currently active primitive $p(k)$, and let $\mathbf{p}_{k+1}$ denote the request point selected for obtaining $p(k+1)$. 
    
        Let $D_{\text{obs}}$ denote the maximum forward perception range of a single visual observation. 
        For this reference analysis, we assume that the nominal path from $\mathbf{p}_k$ to $\mathbf{w}(k)$ lies within the observation's spatial support and has length at most $D_{\text{obs}}$. 
        Travel times are given by reference path lengths divided by $v_{\text{max}}$, without additional turning or stopping delays. 
        Therefore,
        \begin{align}
            \tau_{\text{move}}(\mathbf{p}_k,\mathbf{w}(k))\le \frac{D_{\text{obs}}}{v_{\text{max}}}.
            \label{eq_max_move_time}
        \end{align}
        Here, $D_{\text{obs}}/v_{\text{max}}$ bounds the motion time along the reference path supported by the observation. 
    
        We consider reference continuations through $\mathbf{p}_{k+1}$ whose cumulative detour, measured from the submission location $\mathbf{p}_k$, satisfies 
        \begin{align}
            \tau_{\text{move}}&(\mathbf{p}_k,\mathbf{p}_{k+1}) + \tau_{\text{move}}(\mathbf{p}_{k+1},\mathbf{w}(k)) \le \nonumber \\
            &\tau_{\text{move}}(\mathbf{p}_k,\mathbf{w}(k)) + \varepsilon_{\text{det}}.
            \label{eq_pre_request_detour}
        \end{align}
        Let $\mathbf{q}$ be a communication-feasible retrieval point on the reference continuation from $\mathbf{p}_{k+1}$ to $\mathbf{w}(k)$. 
        We assume that travel times along this continuation are additive. 
        The downstream feasibility condition requires at least $\widehat{T}_{\text{window}}(k+1,\mathbf{p}_{k+1})+\varepsilon_{\text{safe}}$ for travel from $\mathbf{p}_{k+1}$ to $\mathbf{q}$. 
        Since $\mathbf{q}$ precedes $\mathbf{w}(k)$ along the continuation, the remaining travel time to the waypoint satisfies 
        \begin{align}
            \tau_{\text{move}}(\mathbf{p}_{k+1},\mathbf{w}(k)) \ge \widehat{T}_{\text{window}}(k+1,\mathbf{p}_{k+1}) + \varepsilon_{\text{safe}}.
            \label{eq_pre_request_window}
        \end{align}
        Substituting \eqref{eq_pre_request_window} into \eqref{eq_pre_request_detour} yields
        \begin{align}
            \tau_{\text{move}}(\mathbf{p}_k,\mathbf{p}_{k+1}) \le &\; \tau_{\text{move}}(\mathbf{p}_k,\mathbf{w}(k)) \nonumber \\ 
            &- \widehat{T}_{\text{window}}(k+1,\mathbf{p}_{k+1}) - \varepsilon.
            \label{eq_pre_request_time_bound}
        \end{align}
        where $\varepsilon = \varepsilon_{\text{safe}} - \varepsilon_{\text{det}}$ is the signed difference between the retrieval safety margin and the detour allowance. 
        The detour allowance increases the permitted travel time to the current waypoint, whereas the retrieval safety margin reserves time for response retrieval. 
        Consequently, increasing $\varepsilon_{\text{det}}$ relaxes the upper bound on request deferral time, while increasing $\varepsilon_{\text{safe}}$ tightens it. 
        Combining \eqref{eq_max_move_time} and \eqref{eq_pre_request_time_bound}, we obtain
        \begin{align}
            \tau_{\text{move}}(\mathbf{p}_k,\mathbf{p}_{k+1}) \le &\frac{D_{\text{obs}}}{v_{\text{max}}} - \widehat{T}_{\text{window}}(k+1,\mathbf{p}_{k+1}) - \varepsilon.
            \label{eq_limitation_time}
        \end{align}
        \eqref{eq_limitation_time} bounds the time available before the next submission under these reference-motion conditions. 
        Larger communication or inference delays require earlier submission, whereas greater spatial support allows the next request to be deferred farther along the motion. 
        Defining the request deferral length as 
        \begin{align}
            L_{\text{def},k} := v_\text{max} \cdot \tau_{\text{move}}(\mathbf{p}_k,\mathbf{p}_{k+1}),
            \label{eq_pre_request_length}
        \end{align}
        gives 
        \begin{align}
            L_{\text{def},k} \le D_{\text{obs}} - v_\text{max} \left( \widehat{T}_{\text{window}}(k+1,\mathbf{p}_{k+1}) + \varepsilon \right).
            \label{eq_pre_request_length_bound}
        \end{align}
        At the assumed constant speed, $L_{\text{def},k}$ is the reference path length between the two submission locations. 
        \eqref{eq_pre_request_length_bound} allocates the finite observation support between motion before submission and the time reserved for the next cloud cycle. 
        When uplink and downlink transmission times are negligible, the bound reduces to 
        \begin{align}
            L_{\text{def},k}\lesssim D_{\text{obs}} - v_\text{max} \left(\widehat{T}_{\text{infer}}(k+1) + \beta\sigma_{\text{infer}}(k+1) + \varepsilon \right).
        \end{align}
        Cloud inference therefore limits request deferral even under favorable communication conditions. 
        Larger transmission delays further reduce the available distance. 
        A segment without feasible submission locations consumes this deferral budget as the robot traverses it. 
        Continuous reference motion requires a feasible next submission location within the bound, together with a downstream retrieval opportunity. 
        The resulting boundary relates the traversable distance before the next request to observation support, cloud latency, and the two timing margins. 

%% file: references.bib
@article{firoozi2024foundation,
    author = {Roya Firoozi and Johnathan Tucker and Stephen Tian and Anirudha Majumdar and Jiankai Sun and Weiyu Liu and Yuke Zhu and Shuran Song and Ashish Kapoor and Karol Hausman and Brian Ichter and Danny Driess and Jiajun Wu and Cewu Lu and Mac Schwager},
    title = {{Foundation models in robotics: Applications, challenges, and the future}},
    journal = {The International Journal of Robotics Research},
    volume = {44},
    number = {5},
    pages = {701-739},
    year = {2025}
}

@article{zheng2025edge,
    author = {Zheng, Yue and Chen, Yuhao and Qian, Bin and Shi, Xiufang and Shu, Yuanchao and Chen, Jiming},
    title = {{A Review on Edge Large Language Models: Design, Execution, and Applications}},
    year = {2025},
    volume = {57},
    number = {8},
    issn = {0360-0300},
    journal = {ACM Computing Surveys},
    month = mar,
    articleno = {209},
    numpages = {35}
}

@article{pan2026robot,
    author = {Pan, Haotian and Huang, Shibo and Yang, Jian and Mi, Jinpeng and Li, Ke and You, Xiong and Liang, Peidong and Yang, Jinbo and Liu, Yingjie and Zhang, Jianfeng and Wang, Muyu and Yang, Jie and Zhang, Xinyu and Zhao, Lijun and Chen, Mingsong and Zhou, Jie and Wei, Xian},
    title = {{Robot Navigation via Foundation Language Models: A Review}},
    year = {2026},
    volume = {58},
    number = {11},
    issn = {0360-0300},
    journal = {ACM Computing Surveys},
    articleno = {291},
    numpages = {38}
}

@InProceedings{zitkovich23rt2,
    title = 	 {{RT-2: Vision-Language-Action Models Transfer Web Knowledge to Robotic Control}},
    author =       {Zitkovich, Brianna and Yu, Tianhe and Xu, Sichun and Xu, Peng and Xiao, Ted and Xia, Fei and Wu, Jialin and Wohlhart, Paul and Welker, Stefan and Wahid, Ayzaan and Vuong, Quan and Vanhoucke, Vincent and Tran, Huong and Soricut, Radu and Singh, Anikait and Singh, Jaspiar and Sermanet, Pierre and Sanketi, Pannag R. and Salazar, Grecia and Ryoo, Michael S. and Reymann, Krista and Rao, Kanishka and Pertsch, Karl and Mordatch, Igor and Michalewski, Henryk and Lu, Yao and Levine, Sergey and Lee, Lisa and Lee, Tsang-Wei Edward and Leal, Isabel and Kuang, Yuheng and Kalashnikov, Dmitry and Julian, Ryan and Joshi, Nikhil J. and Irpan, Alex and Ichter, Brian and Hsu, Jasmine and Herzog, Alexander and Hausman, Karol and Gopalakrishnan, Keerthana and Fu, Chuyuan and Florence, Pete and Finn, Chelsea and Dubey, Kumar Avinava and Driess, Danny and Ding, Tianli and Choromanski, Krzysztof Marcin and Chen, Xi and Chebotar, Yevgen and Carbajal, Justice and Brown, Noah and Brohan, Anthony and Arenas, Montserrat Gonzalez and Han, Kehang},
    booktitle = 	 {Proceedings of the 7th Annual Conference on Robot Learning (CoRL)},
    pages = 	 {2165--2183},
    year = 	 {2023},
    volume = 	 {229}
}

@article{kim2024openvla,
    title={{OpenVLA: An Open-Source Vision-Language-Action Model}}, 
    author={{Moo Jin} Kim and Karl Pertsch and Siddharth Karamcheti and Ted Xiao and Ashwin Balakrishna and Suraj Nair and Rafael Rafailov and Ethan Foster and Grace Lam and Pannag Sanketi and Quan Vuong and Thomas Kollar and Benjamin Burchfiel and Russ Tedrake and Dorsa Sadigh and Sergey Levine and Percy Liang and Chelsea Finn},
    year={2024},
    eprint={2406.09246},
    journal = {arXiv preprint arXiv:2406.09246}
}

@article{black2026pi0,
    title={$\pi_0$: {A Vision-Language-Action Flow Model for General Robot Control}}, 
    author={Kevin Black and Noah Brown and Danny Driess and Adnan Esmail and Michael Equi and Chelsea Finn and Niccolo Fusai and Lachy Groom and Karol Hausman and Brian Ichter and Szymon Jakubczak and Tim Jones and Liyiming Ke and Sergey Levine and Adrian Li-Bell and Mohith Mothukuri and Suraj Nair and Karl Pertsch and Lucy Xiaoyang Shi and James Tanner and Quan Vuong and Anna Walling and Haohuan Wang and Ury Zhilinsky},
    year={2024},
    eprint={2410.24164},
    journal = {arXiv preprint arXiv: 2410.24164}
}

@inproceedings{shah2022lmnav, 
    title={{LM-Nav: Robotic Navigation with Large Pre-Trained Models of Language, Vision, and Action}}, 
    author={Dhruv Shah and Blazej Osinski and Brian Ichter and Sergey Levine}, 
    booktitle={Proceedings of the 6th Annual Conference on Robot Learning (CoRL)}, 
    year={2022} 
}

@ARTICLE{obata2025lipllm, 
    author={Obata, Kazuma and Aoki, Tatsuya and Horii, Takato and Taniguchi, Tadahiro and Nagai, Takayuki}, 
    journal={IEEE Robotics and Automation Letters}, 
    title={{LiP-LLM: Integrating Linear Programming and Dependency Graph With Large Language Models for Multi-Robot Task Planning}}, 
    year={2025}, 
    volume={10}, 
    number={2}, 
    pages={1122-1129}
}

@ARTICLE{wang2025boosting, 
    author={Wang, Jiawei and Wang, Teng and Cai, Wenzhe and Xu, Lele and Sun, Changyin}, 
    journal={IEEE Robotics and Automation Letters}, 
    title={{Boosting Efficient Reinforcement Learning for Vision-and-Language Navigation With Open-Sourced LLM}}, 
    year={2025}, 
    volume={10}, 
    number={1}, 
    pages={612-619}
}

@INPROCEEDINGS{liang2023code,
    author={Liang, Jacky and Huang, Wenlong and Xia, Fei and Xu, Peng and Hausman, Karol and Ichter, Brian and Florence, Pete and Zeng, Andy},
    booktitle={Proceedings of the 2023 IEEE International Conference on Robotics and Automation (ICRA)}, 
    title={{Code as Policies: Language Model Programs for Embodied Control}}, 
    year={2023},
    volume={},
    number={},
    pages={9493-9500}
}

@INPROCEEDINGS{huang2023visual,
    author={Huang, Chenguang and Mees, Oier and Zeng, Andy and Burgard, Wolfram},
    booktitle={Proceedings of the 2023 IEEE International Conference on Robotics and Automation (ICRA)}, 
    title={{Visual Language Maps for Robot Navigation}}, 
    year={2023},
    volume={},
    number={},
    pages={10608-10615}
}

@article{liu2026event,
    title={{Event-Driven Proactive Assistive Manipulation with Grounded Vision-Language Planning}}, 
    author={Fengkai Liu and Hao Su and Haozhuang Chi and Rui Geng and Congzhi Ren and Xuqing Liu and Yucheng Xu and Yuichi Ohsita and Liyun Zhang},
    year={2026},
    eprint={2603.23950},
    journal = {arXiv preprint arXiv: 2603.23950},
}

@ARTICLE{zeydan2025role,
    author={Zeydan, Engin and Arslan, Suayb S. and Turk, Yekta and Hewa, Tharaka and Liyanage, Madhusanka},
    journal={IEEE Open Journal of the Communications Society}, 
    title={{The Role of Mobile Communications for Industrial Automation: Architecture, Applications and Challenges}}, 
    year={2025},
    volume={6},
    number={},
    pages={6808-6841}
}

@INPROCEEDINGS{ichnowski2023fogros2,
    author={Ichnowski, Jeffrey and Chen, Kaiyuan and Dharmarajan, Karthik and Adebola, Simeon and Danielczuk, Michael and Mayoral-Vilches, Víctor and Jha, Nikhil and Zhan, Hugo and Llontop, Edith and Xu, Derek and Buscaron, Camilo and Kubiatowicz, John and Stoica, Ion and Gonzalez, Joseph and Goldberg, Ken},
    booktitle={Proceedings of the 2023 IEEE International Conference on Robotics and Automation (ICRA)}, 
    title={{FogROS2: An Adaptive Platform for Cloud and Fog Robotics Using ROS 2}}, 
    year={2023},
    volume={},
    number={},
    pages={5493-5500}
}

@INPROCEEDINGS{chen2024fogros2ls,
    author={Chen, Kaiyuan and Wang, Michael and Gualtieri, Marcus and Tian, Nan and Juette, Christian and Ren, Liu and Ichnowski, Jeffrey and Kubiatowicz, John and Goldberg, Ken},
    booktitle={Proceedings of the 2024 IEEE International Conference on Robotics and Automation (ICRA)}, 
    title={{FogROS2-LS: A Location-Independent Fog Robotics Framework for Latency Sensitive ROS2 Applications}}, 
    year={2024},
    volume={},
    number={},
    pages={10581-10587}
}

@INPROCEEDINGS{chen2024fogros2ft,
    author={Chen, Kaiyuan and Hari, Kush and Chung, Trinity and Wang, Michael and Tian, Nan and Juette, Christian and Ichnowski, Jeffrey and Ren, Liu and Kubiatowicz, John and Stoica, Ion and Goldberg, Ken},
    booktitle={Proceedings of the 2024 IEEE/RSJ International Conference on Intelligent Robots and Systems (IROS)}, 
    title={{FogROS2-FT: Fault Tolerant Cloud Robotics}}, 
    year={2024},
    volume={},
    number={},
    pages={1390-1397}
}

@INPROCEEDINGS{chen2025fogros2plr,
    author={Chen, Kaiyuan and Tian, Nan and Juette, Christian and Qiu, Tianshuang and Ren, Liu and Kubiatowicz, John and Goldberg, Ken},
    booktitle={Proceedings of the 2025 IEEE International Conference on Robotics and Automation (ICRA)}, 
    title={{FogROS2-PLR: Probabilistic Latency-Reliability for Cloud Robotics}}, 
    year={2025},
    volume={},
    number={},
    pages={16290-16297}
}

@article{tang2025vlash,
    title={{VLASH: Real-Time VLAs via Future-State-Aware Asynchronous Inference}}, 
    author={Jiaming Tang and Yufei Sun and Yilong Zhao and Shang Yang and Yujun Lin and Zhuoyang Zhang and James Hou and Yao Lu and Zhijian Liu and Song Han},
    year={2025},
    eprint={2512.01031},
    archivePrefix={arXiv},
    primaryClass={cs.RO},
    journal = {arXiv preprint arXiv: 2512.01031},
}

@article{hirose2026asyncvla,
    title={{AsyncVLA: An Asynchronous VLA for Fast and Robust Navigation on the Edge}}, 
    author={Noriaki Hirose and Catherine Glossop and Dhruv Shah and Sergey Levine},
    year={2026},
    eprint={2602.13476},
    archivePrefix={arXiv},
    primaryClass={cs.RO},
    journal = {arXiv preprint arXiv: 2602.13476}
}

@article{nguyen2026speculative,
    title={{Speculative Policy Orchestration: A Latency-Resilient Framework for Cloud-Robotic Manipulation}}, 
    author={Chanh Nguyen and Shutong Jin and Florian T. Pokorny and Erik Elmroth},
    year={2026},
    eprint={2603.19418},
    journal = {arXiv preprint arXiv: 2603.19418}
}

@article{tariq2025robust,
    title = {Robust mobile robot path planning via LLM-based dynamic waypoint generation},
    journal = {Expert Systems with Applications},
    volume = {282},
    pages = {127600},
    year = {2025},
    issn = {0957-4174},
    author = {Muhammad Taha Tariq and Yasir Hussain and Congqing Wang}
}

@ARTICLE{he2024large,
    author={He, Ying and Fang, Jingcheng and Yu, F. Richard and Leung, Victor C.},
    journal={IEEE Transactions on Mobile Computing}, 
    title={{Large Language Models (LLMs) Inference Offloading and Resource Allocation in Cloud-Edge Computing: An Active Inference Approach}}, 
    year={2024},
    volume={23},
    number={12},
    pages={11253-11264}
}

@article{zheng2026rapid,
    title={{RAPID: Redundancy-Aware and Compatibility-Optimal Edge-Cloud Partitioned Inference for Diverse VLA Models}}, 
    author={Zihao Zheng and Sicheng Tian and Hangyu Cao and Chenyue Li and Jiayu Chen and Maoliang Li and Xinhao Sun and Hailong Zou and Guojie Luo and Xiang Chen},
    year={2026},
    eprint={2603.07949},
    journal = {arXiv preprint arXiv: 2603.07949}
}

@ARTICLE{qu2025mobile,
    author={Qu, Guanqiao and Chen, Qiyuan and Wei, Wei and Lin, Zheng and Chen, Xianhao and Huang, Kaibin},
    journal={IEEE Communications Surveys \& Tutorials}, 
    title={{Mobile Edge Intelligence for Large Language Models: A Contemporary Survey}}, 
    year={2025},
    volume={27},
    number={6},
    pages={3820-3860}
}

@ARTICLE{jin2026moe2,
    author={Jin, Lyudong and Zhang, Yanning and Li, Yanhan and Wang, Shurong and Yang, Howard H. and Wu, Jian and Zhang, Meng},
    journal={IEEE Transactions on Networking}, 
    title={{MoE2: Optimizing Collaborative Inference for Edge Large Language Models}}, 
    year={2026},
    volume={34},
    number={},
    pages={4637-4651}
}

@inproceedings{yuan2025local,
    author = {Yuan, Liangqi and Han, Dong-Jun and Wang, Shiqiang and Brinton, Christopher},
    title = {{Local-Cloud Inference Offloading for LLMs in Multi-Modal, Multi-Task, Multi-Dialogue Settings}},
    year = {2025},
    isbn = {9798400713538},
    pages = {201–210},
    numpages = {10},
    booktitle = {Proceedings of the 26th International Symposium on Theory, Algorithmic Foundations, and Protocol Design for Mobile Networks and Mobile Computing},
}

@article{yang2026asyncshield,
    title={{AsyncShield: A Plug-and-Play Edge Adapter for Asynchronous Cloud-based VLA Navigation}}, 
    author={Kai Yang and Zedong Chu and Yingnan Guo and Zhengbo Wang and Shichao Xie and Yanfen Shen and Xiaolong Wu and Xing Li and Mu Xu},
    year={2026},
    eprint={2604.24086},
    journal = {arXiv preprint arXiv: 2604.24086}
}

@ARTICLE{ghaffarkhah2011communication,
    author={Ghaffarkhah, Alireza and Mostofi, Yasamin},
    journal={IEEE Transactions on Automatic Control}, 
    title={{Communication-Aware Motion Planning in Mobile Networks}}, 
    year={2011},
    volume={56},
    number={10},
    pages={2478-2485}
}

@article{twigg2013efficient,
    title={{Efficient base station connectivity area discovery}},
    author={Twigg, Jeffrey N and Fink, Jonathan R and Yu, Paul L and Sadler, Brian M},
    journal={The International Journal of Robotics Research},
    volume={32},
    number={12},
    pages={1398--1410},
    year={2013}
}

@INPROCEEDINGS{caccamo2017rcamp,
    author={Caccamo, Sergio and Parasuraman, Ramviyas and Freda, Luigi and Gianni, Mario and Ögren, Petter},
    booktitle={Proceedings of the 2017 IEEE/RSJ International Conference on Intelligent Robots and Systems (IROS)}, 
    title={{RCAMP: A resilient communication-aware motion planner for mobile robots with autonomous repair of wireless connectivity}}, 
    year={2017},
    volume={},
    number={},
    pages={2010-2017}
}

@ARTICLE{zavlanos2009distributed,
    author={Zavlanos, Michael M. and Pappas, George J.},
    journal={IEEE Transactions on Robotics}, 
    title={{Distributed Connectivity Control of Mobile Networks}}, 
    year={2008},
    volume={24},
    number={6},
    pages={1416-1428}
}

@INPROCEEDINGS{yang2019connectivity,
    author={Yang, Hongyu and Zhang, Jun and Song, S.H. and Lataief, Khaled B.},
    booktitle={Proceedings of the 2019 IEEE Wireless Communications and Networking Conference (WCNC)}, 
    title={{Connectivity-Aware UAV Path Planning with Aerial Coverage Maps}}, 
    year={2019},
    volume={},
    number={},
    pages={1-6}
}

@INPROCEEDINGS{yang2024integrating,
    author={Yang, Yupeng and Lyu, Yiwei and Zhang, Yanze and Gao, Ian and Luo, Wenhao},
    booktitle={Proceedings of the 2024 IEEE/RSJ International Conference on Intelligent Robots and Systems (IROS)}, 
    title={{Integrating Online Learning and Connectivity Maintenance for Communication-Aware Multi-Robot Coordination}}, 
    year={2024},
    volume={},
    number={},
    pages={5770-5776}
}

@ARTICLE{kantaros2017distributed,
  author={Kantaros, Yiannis and Zavlanos, Michael M.},
  journal={IEEE Transactions on Automatic Control}, 
  title={{Distributed Intermittent Connectivity Control of Mobile Robot Networks}}, 
  year={2017},
  volume={62},
  number={7},
  pages={3109-3121}
}

@ARTICLE{kantaros2019temporal,
    author={Kantaros, Yiannis and Guo, Meng and Zavlanos, Michael M.},
    journal={IEEE Transactions on Automatic Control}, 
    title={{Temporal Logic Task Planning and Intermittent Connectivity Control of Mobile Robot Networks}}, 
    year={2019},
    volume={64},
    number={10},
    pages={4105-4120}
}

@ARTICLE{li2025edge,
  author={Li, Guoliang and Han, Ruihua and Wang, Shuai and Gao, Fei and Eldar, Yonina C. and Xu, Chengzhong},
  journal={IEEE/ASME Transactions on Mechatronics}, 
  title={{Edge Accelerated Robot Navigation With Collaborative Motion Planning}}, 
  year={2025},
  volume={30},
  number={2},
  pages={1166-1178}
}

@INPROCEEDINGS{clark2022propeml, 
    AUTHOR    = {Lillian Clark AND Jeffrey Edlund AND {Marc Sanchez} Net AND {Tiago Stegun} Vaquero AND Ali-akbar Agha-mohammadi}, 
    TITLE     = {{PropEM-L: Radio Propagation Environment Modeling and Learning for Communication-Aware Multi-Robot Exploration}}, 
    BOOKTITLE = {Proceedings of Robotics: Science and Systems (RSS)}, 
    YEAR      = {2022}
}

@ARTICLE{hurst2023optimization,
    author={Hurst, Winston and Mostofi, Yasamin},
    journal={IEEE Transactions on Wireless Communications}, 
    title={{Optimization of Mobile Robotic Relay Operation for Minimal Average Wait Time}}, 
    year={2023},
    volume={22},
    number={6},
    pages={3733-3747}
}

@ARTICLE{yan2013co,
    author={Yan, Yuan and Mostofi, Yasamin},
    journal={IEEE Transactions on Wireless Communications}, 
    title={{Co-Optimization of Communication and Motion Planning of a Robotic Operation under Resource Constraints and in Fading Environments}}, 
    year={2013},
    volume={12},
    number={4},
    pages={1562-1572}
}

@INPROCEEDINGS{muralidharan2017path,
    author={Muralidharan, Arjun and Mostofi, Yasamin},
    booktitle={Proceedings of the 2017 IEEE Globecom Workshops (GC Wkshps)}, 
    title={{Path Planning for a Connectivity Seeking Robot}}, 
    year={2017},
    volume={},
    number={},
    pages={1-6}
}

@ARTICLE{ali2019motion,
    author={Ali, Usman and Cai, Hong and Mostofi, Yasamin and Wardi, Yorai},
    journal={IEEE Transactions on Control of Network Systems}, 
    title={{Motion-Communication Co-Optimization With Cooperative Load Transfer in Mobile Robotics: An Optimal Control Perspective}}, 
    year={2019},
    volume={6},
    number={2},
    pages={621-632}
}

@article{gil2015adaptive,
    title={{Adaptive communication in multi-robot systems using directionality of signal strength}},
    author={Gil, Stephanie and Kumar, Swarun and Katabi, Dina and Rus, Daniela},
    journal={The International Journal of Robotics Research},
    volume={34},
    number={7},
    pages={946--968},
    year={2015}
}

@article{achey2026rf,
    title={{RF-Modulated Adaptive Communication Improves Multi-Agent Robotic Exploration}}, 
    author={Lorin Achey and Breanne Crockett and Christoffer Heckman and Bradley Hayes},
    year={2026},
    eprint={2602.12074},
    journal = {arXiv preprint arXiv: 2602.12074}
}

@article{parwez2026ctmap,
    title={{CTMap: LLM-Enabled Connectivity-Aware Path Planning in Millimeter-Wave Digital Twin Networks}}, 
    author={Md Salik Parwez and Sai Teja Srivillibhutturu and Sai Venkat Reddy Kopparthi and Asfiya Misba and Debashri Roy and Habeeb Olufowobi and Charles Kim},
    year={2026},
    eprint={2601.00110},
    journal = {arXiv preprint arXiv: 2601.00110}
}

@INPROCEEDINGS{psomiadis2024communication,
    author={Psomiadis, Evangelos and Maity, Dipankar and Tsiotras, Panagiotis},
    booktitle={Proceedings of the 2024 IEEE International Conference on Robotics and Automation (ICRA)}, 
    title={{Communication-Aware Map Compression for Online Path-Planning}}, 
    year={2024},
    volume={},
    number={},
    pages={12368-12374}
}

@article{bi2022supplementary,
    title={{Supplementary open dataset for WiFi indoor localization based on received signal strength}},
    author={Bi, Jingxue and Wang, Yunjia and Yu, Baoguo and Cao, Hongji and Shi, Tongguang and Huang, Lu},
    journal={Satellite Navigation},
    volume={3},
    number={1},
    pages={25},
    year={2022},
    publisher={Springer}
}

@article{ruan2026latency,
    title={{A Latency-Aware Framework for Visuomotor Policy Learning on Industrial Robots}}, 
    author={Daniel Ruan and Salma Mozaffari and Sigrid Adriaenssens and Arash Adel},
    year={2026},
    eprint={2602.14255},
    journal = {arXiv preprint arXiv: 2602.14255}
}

@inproceedings{wei2026ground,
    title={{Ground Slow, Move Fast: A Dual-System Foundation Model for Generalizable Vision-Language Navigation}},
    author={Meng Wei and Chenyang Wan and Jiaqi Peng and Xiqian Yu and Yuqiang Yang and Delin Feng and Wenzhe Cai and Chenming Zhu and Tai Wang and Jiangmiao Pang and Xihui Liu},
    booktitle={Proceedings of the 14th International Conference on Learning Representations (ICLR)},
    year={2026}
}

@inproceedings{anderson2018vision,
  title={Vision-and-language navigation: Interpreting visually-grounded navigation instructions in real environments},
  author={Anderson, Peter and Wu, Qi and Teney, Damien and Bruce, Jake and Johnson, Mark and S{\"u}nderhauf, Niko and Reid, Ian and Gould, Stephen and Van Den Hengel, Anton},
  booktitle={Proceedings of the 2018 IEEE/CVF Conference on Computer Vision and Pattern Recognition (CVPR)},
  pages={3674--3683},
  year={2018}
}

@inproceedings{zhou2024navgpt,
  title={Navgpt: Explicit reasoning in vision-and-language navigation with large language models},
  author={Zhou, Gengze and Hong, Yicong and Wu, Qi},
  booktitle={Proceedings of the AAAI Conference on Artificial Intelligence},
  volume={38},
  number={7},
  pages={7641--7649},
  year={2024}
}

@INPROCEEDINGS{hurst2021communication,
  author={Hurst, Winston and Cai, Hong and Mostofi, Yasamin},
  booktitle={Proceedings of the IEEE International Conference on Communications (ICC)}, 
  title={{Communication-Aware RRT*: Path Planning for Robotic Communication Operation in Obstacle Environments}}, 
  year={2021},
  pages={1-6}
}

@ARTICLE{afzal2023model,
  author={Afzal, Waqas and Masoud, Ahmad A.},
  journal={IEEE Transactions on Automation Science and Engineering}, 
  title={{Model-Based Navigation Control for Communication-Aware Motion Using Harmonic Potential Fields}}, 
  year={2023},
  volume={20},
  number={4},
  pages={2636-2654}
}

@article{kim2026comp,
    author = {Kim, Mingi and Kim, Hyungki},
    title = {{Comp-PointLLM: An explainable multimodal large language model for 3D object comparison}},
    journal = {Journal of Computational Design and Engineering},
    volume = {13},
    number = {5},
    pages = {122-147},
    year = {2026},
    month = {05}
}

@article{yun2025smartlab,
    author = {Yun, Won-Jae and Hong, Sungjin and Lee, Kyu-Wha and Kim, Dohyeon and Kim, Hyungjung and Ahn, Sung-Hoon},
    title = {SmartLab: Flexible and interoperable manufacturing laboratory system for remote education and research using a mobile manipulator},
    journal = {Journal of Computational Design and Engineering},
    volume = {12},
    number = {4},
    pages = {221-235},
    year = {2025},
    month = {04}
}
